\documentclass{article} 
\usepackage{colm2025_conference}

\usepackage[expansion=false]{microtype}
\usepackage{graphicx}
\usepackage{subcaption}
\usepackage{booktabs} 

\usepackage{hyperref}
\usepackage{url}

\usepackage{amsmath}
\usepackage{amssymb}
\usepackage{mathtools}
\usepackage{amsthm}
\usepackage[normalem]{ulem}

\usepackage[capitalize,noabbrev]{cleveref}

\usepackage{multirow}
\usepackage[table]{xcolor}
\usepackage[most]{tcolorbox}
\tcbuselibrary{skins,breakable}
\usepackage{xspace}
\usepackage{enumitem}
\usepackage{colortbl}
\usepackage{siunitx}    
\usepackage{makecell,threeparttable}
\crefname{section}{Section}{Sections}
\Crefname{section}{Section}{Sections}
\crefrangelabelformat{section}{#3#1--#4#2}

\usepackage{amsfonts}
\usepackage{mathrsfs}

\newcommand\eat[1]{}

\usepackage{algorithm}
\usepackage{algorithmic}

\usepackage{amsmath,amsfonts,bm}

\def\eqref#1{equation~\ref{#1}}

\def\1{\bm{1}}

\DeclareMathAlphabet{\mathsfit}{\encodingdefault}{\sfdefault}{m}{sl}
\SetMathAlphabet{\mathsfit}{bold}{\encodingdefault}{\sfdefault}{bx}{n}

\theoremstyle{plain}
\newtheorem{theorem}{Theorem}[section]

\theoremstyle{definition}
\newtheorem{definition}[theorem]{Definition}

\theoremstyle{remark}

\usepackage{listings}
\usepackage{wrapfig}
\usepackage{verbatim} 
\usepackage{tabularx} 
\usepackage{arydshln}

\title{Voting with the Graph: Stable RLAIF via Topological Consistency Maximization}

\author{%
  Boyin Liu$^{1}$ \quad Zhuo Zhang$^{1}$ \quad Sen Huang$^{1}$ \quad Lipeng Xie$^{1}$ \quad Qingxu Fu$^{1}$ \quad Haoran Chen$^{1}$ \\
  Li Yu$^{1}$ \quad Tianyi Hu$^{1}$ \quad Zhaoyang Liu$^{1*}$ \quad Bolin Ding$^{1}$ \quad Dongbin Zhao$^{2}$ \\[4pt]
  $^{1}$Alibaba Group \qquad $^{2}$Chinese Academy of Sciences Institute of Automation \\[2pt]
  {\small \texttt{\{liuboyin.lby, zz297429, huangsen.huang, xielipeng.xlp, fuqingxu.fqx, congling.chr,}} \\
  {\small \texttt{liyu.ly, hutianyi.hty, jingmu.lzy, bolin.ding\}@alibaba-inc.com, dongbin.zhao@ia.ac.cn}}
}

\begin{document}

\maketitle

\begin{abstract}
Reinforcement Learning from AI Feedback (RLAIF) relies on LLM judges as preference measurement instruments, yet these instruments are fundamentally limited by \textbf{random measurement errors}---stochastic fluctuations that manifest as preference cycles (e.g., $A \succ B \succ C \succ A$), occurring in 5--9\% of evaluations across state-of-the-art models. While repeated sampling mitigates noise by averaging multiple judgments, it treats each comparison in isolation and fails to exploit the structural constraints that distinguish systematic signals from random noise. We introduce \textbf{Topological Consensus Rewards (TCR)}, a framework that leverages transitivity as a denoising mechanism via topological majority voting: systematic signals reinforce each other through transitive chains, while random errors cluster into topologically exposed cycles. TCR approximates the Maximum Acyclic Subgraph to filter stochastic noise from preference signals. We also propose \textbf{Cycle Incidence Rate (CIR)} as a diagnostic metric that measures the proportion of samples containing preference cycles. Under our noise model, these cycles arise primarily from stochastic measurement errors rather than genuine intransitivity. Experiments on Arena-Hard, MT-Bench, and WritingBench demonstrate that TCR consistently outperforms pairwise baselines and classical ranking algorithms, while exhibiting robust performance across different judge models.\end{abstract}

\section{Introduction}

Reinforcement Learning from AI Feedback (RLAIF) \citep{bai2022constitutional, lee2023rlaif} has emerged as a scalable alternative to costly human annotation for aligning large language models. Within RLAIF, the pairwise comparison paradigm—where an LLM judge selects the better of two responses—has become the de facto standard, now a core component in aligning leading models such as Kimi K2 \citep{team2025kimi} and NVIDIA Nemotron 3 \citep{nvidia2025nemotron3}.
Unlike pointwise scoring, pairwise comparisons excel at capturing fine-grained preference signals \citep{xu2025pairwise}. \textbf{However, while pairwise comparisons offer finer-grained judgment, their effectiveness for policy optimization depends on their structural consistency.}

\begin{figure}[t]
\centering
\begin{subfigure}[t]{0.49\columnwidth}
\centering
\includegraphics[width=\columnwidth]{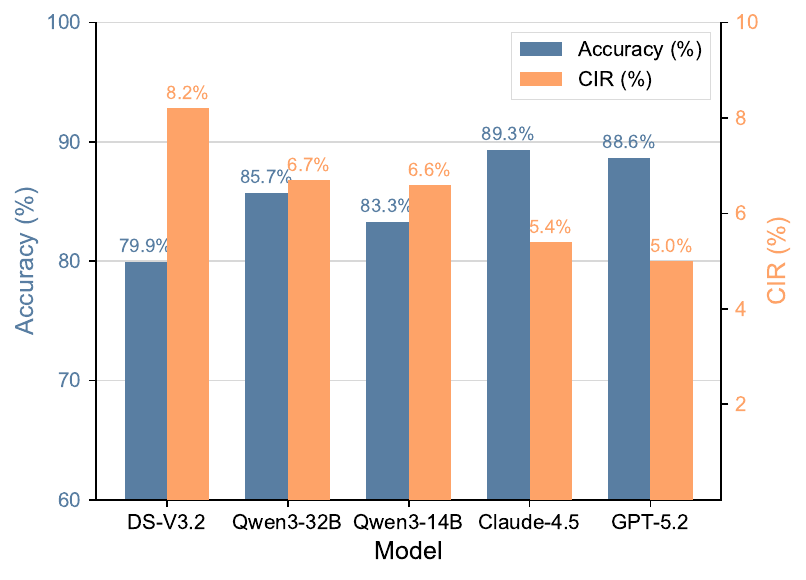}
\caption{}
\label{fig:conflict_rates_models}
\end{subfigure}
\hfill
\begin{subfigure}[t]{0.49\columnwidth}
\centering
\includegraphics[width=\columnwidth]{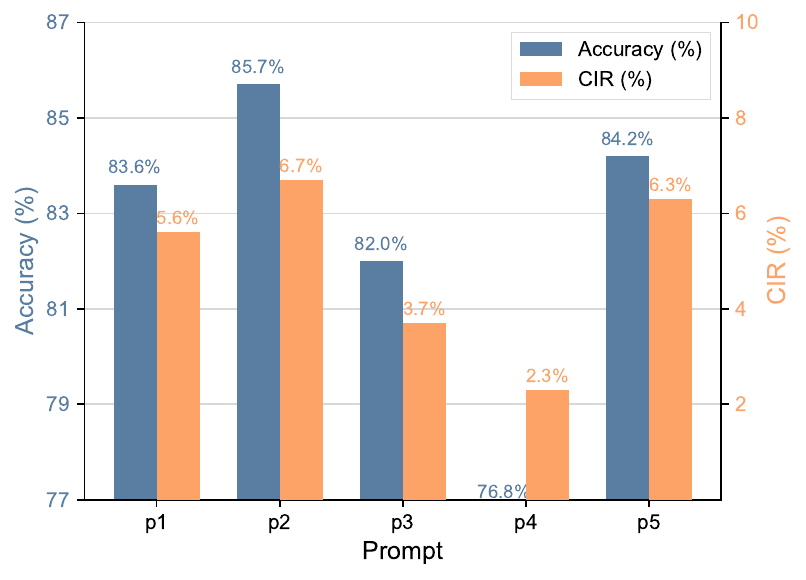}
\caption{}
\label{fig:prompt_optimization}
\end{subfigure}
\caption{Cycle incidence analysis of LLM judges. (a) CIR (Cycle Incidence Rate) and accuracy comparison across models (model versions in Appendix~\ref{sec:model_versions}). (b) Impact of prompt engineering on CIR for Qwen3-32B. Prompts P1-P5 are detailed in Appendix~\ref{sec:pairwise_prompts}. All metrics computed on RewardBench2 \citep{malik2025rewardbench} with $G=4$ candidates. Accuracy excludes tied outputs; CIR is computed on all samples with average edge density $\approx 0.92$ (i.e., 92\% of pairwise comparisons yield non-tie judgments).}
\label{fig:cycle_analysis}
\end{figure}

From a measurement theory perspective, the errors of any instrument can be decomposed into \textit{systematic errors} (consistent biases) and \textit{random errors} (stochastic fluctuations). Currently, the quality of AI judges is evaluated primarily through \textbf{accuracy} on isolated comparisons, which captures aggregate performance but overlooks \textbf{stochastic stability}. An LLM judge may act as a noisy instrument, producing contradictory preferences like $A \succ B \succ C \succ A$ due to random fluctuations. As shown in Figure~\ref{fig:cycle_analysis}(a), such cycles---the topological manifestation of random measurement errors---are non-negligible, occurring in 5--9\% of evaluations across state-of-the-art models (e.g., 5.0\% for GPT-5.2, up to 8.2\% for DeepSeek-V3.2).
These are not merely local inconsistencies; they represent \textbf{stochastic noise} that violates the fundamental topological axioms of ranking, injecting contradictory gradients into policy optimization. Repeated sampling approaches can mitigate noise by averaging multiple judgments for the same comparison pair. However, they treat each comparison in isolation, failing to exploit the structural constraints inherent in preference graphs. In contrast, transitivity offers a form of \textit{structural self-consistency}: it allows the system to cross-validate edges against one another (e.g., $A \to B$ and $B \to C$ validate $A \to C$), providing a more rigorous mechanism for noise elimination than simple statistical averaging.

We argue that preference cycles are diagnostic signals that localize random errors. Unlike systematic errors which are consistent (e.g., always preferring longer answers), random errors disrupt the graph's topology by violating transitivity. Just as statistical averaging reduces random error in scalar measurements, we propose a \textit{topological majority voting} mechanism that filters stochastic noise by enforcing topological consistency. Transitivity serves as a denoising mechanism: systematic signals $A \to B$ and $B \to C$ structurally ``vote'' for $A \to C$, while random errors cluster into topologically exposed cycles. By analyzing the preference graph globally, we allow the coherent systematic signal to effectively ``outvote'' the stochastic noise.

To quantify this phenomenon, we introduce the \textbf{Cycle Incidence Rate (CIR)} as a diagnostic metric that measures the proportion of samples containing preference cycles---the topological signature of random measurement errors under our noise model (Appendix~\ref{sec:theoretical_foundation}). This reveals that accuracy and stochastic stability are distinct dimensions of judge quality that do not always align (Figure~\ref{fig:cycle_analysis}(b)). Building on this insight, we propose \textbf{Topological Consensus Rewards (TCR)}, which approximates the \textit{Maximum Acyclic Subgraph (MAS)}—the largest subset of edges consistent with a single global ranking—via an efficient greedy heuristic, filtering stochastic noise from raw preference signals and producing a purified reward that integrates with existing optimizers like GRPO \citep{wang2025pref}.

In summary, our contributions are:
\begin{itemize}[leftmargin=*, nosep]
    \item \textbf{Metric:} We introduce the Cycle Incidence Rate (CIR), a label-free diagnostic metric that quantifies the stochastic stability of LLM judges, revealing that stability is an orthogonal dimension to accuracy.
    \item \textbf{Method:} We propose Topological Consensus Rewards (TCR), a plug-and-play framework that filters random measurement errors by approximating the Maximum Acyclic Subgraph (MAS), enabling coherent reward computation from noisy judgments.
    \item \textbf{Validation:} Extensive experiments on Arena-Hard, MT-Bench, and WritingBench demonstrate that TCR consistently outperforms standard pairwise methods and learned reward models, providing a robust enhancement for RLAIF.
\end{itemize}

\section{Related Work}

\paragraph{RLAIF and Preference Learning.}
The pairwise comparison paradigm is widely adopted in RLAIF \citep{bai2022constitutional, lee2023rlaif, ouyang2022training, rafailov2024direct}.
However, its efficacy is limited by LLM judge measurement errors, specifically random errors that produce preference cycles violating transitivity. While prior work has framed logical consistency conceptually \citep{liu2024aligning}, the community lacks a formal metric to quantify these errors. We address this by introducing the \textbf{Cycle Incidence Rate (CIR)}.

\paragraph{Preference Aggregation Methods.}
Classical methods for aggregating pairwise preferences are well-established in the ranking literature, including Bradley-Terry-Luce models \citep{bradley1952rank, hunter2004mm}, Rank Centrality \citep{negahban2012iterative}, HodgeRank \citep{jiang2011statistical}, and Kemeny ranking \citep{ailon2008aggregating}. However, these methods were designed for \textit{offline} ranking tasks where the goal is to produce a single global ordering. Their application to \textit{online} reinforcement learning—where preferences must be repeatedly aggregated for each batch of sampled responses—remains unexplored. Our work bridges this gap by integrating consensus extraction directly into the RL training loop, with theoretical guarantees (Appendix~\ref{sec:theoretical_foundation}) validated through Monte Carlo experiments.

\paragraph{Random Error Filtering in RLAIF.}
Existing approaches to preference inconsistencies treat them as statistical noise, resorting to data selection \citep{deng2025less} or implicit handling through architecture design \citep{wang2025gram}. These methods do not explicitly filter the random errors that produce preference cycles. In contrast, our \textbf{Topological Consensus Rewards (TCR)} framework is a modular denoising filter that approximates the Maximum Acyclic Subgraph to filter stochastic noise and produce globally coherent rewards \textit{before} policy optimization. This plug-and-play design allows TCR to enhance any optimizer (e.g., GRPO \citep{wang2025pref}, GSPO \citep{zheng2025group}) without modifying its internals.

\section{Methodology}

We develop our approach in two stages. First, we introduce the \textbf{Cycle Incidence Rate (CIR)}, a diagnostic metric that quantifies a previously overlooked dimension of LLM judge quality: \textit{stochastic stability} (\S\ref{sec:cir}). CIR measures the proportion of samples containing preference cycles, which under our noise model arise primarily from random measurement errors. Our analysis reveals that even state-of-the-art judges exhibit cycle incidence rates up to 6.7\%, and that accuracy and stochastic stability represent \textit{distinct, sometimes opposing} optimization targets.

\begin{figure*}[!t]
\centering
\includegraphics[width=\textwidth]{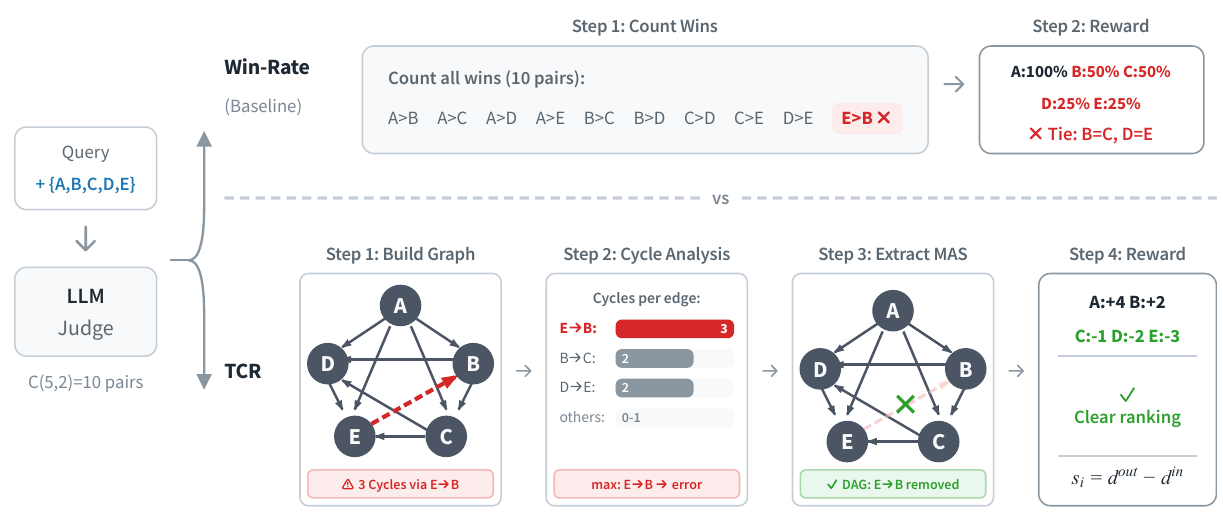}
\caption{\textbf{The Topological Consensus Rewards (TCR) framework versus standard Win-Rate baselines.}
While traditional methods (Top) treat pairwise judgments in isolation, often leading to uninformative ties in the presence of cycles (e.g., $B \to E \to B$ causes $B=C$ at 50\% win-rate), TCR (Bottom) leverages the graph's global topology.
\textbf{(1) Graph Construction:} TCR constructs a directed preference graph from all pairwise judgments.
\textbf{(2) Find Optimal Ordering:} TCR computes a flow score $\delta_i = d_i^{\text{out}} - d_i^{\text{in}}$ for each node and sorts by this score to obtain a candidate global ranking $\pi^*$.
\textbf{(3) Approximate MAS:} Edges consistent with $\pi^*$ (forward edges) are retained; edges violating $\pi^*$ (backward edges, marked with \textcolor{red}{$\times$}, e.g., $E \to B$) are identified as cycle-inducing and excluded by TCR, yielding an acyclic subgraph that approximates the Maximum Acyclic Subgraph.
\textbf{(4) Reward Computation:} The purified DAG yields net-degree scores that induce a ranking (e.g., $A \succ B \succ C \succ D \succ E$; ties possible when nodes have equal scores) for coherent policy optimization.}
\label{fig:tcr_framework}
\end{figure*}

Second, we propose \textbf{Topological Consensus Rewards (TCR)}, which operationalizes our key insight: preference cycles are the topological manifestation of random measurement errors---they localize where stochastic fluctuations disrupt the systematic signal implied by transitivity (\S\ref{sec:tcr}). TCR uses the greedy heuristic from \citet{eades1993fast}---originally designed for the Feedback Arc Set (FAS) problem---to identify and remove cycle-inducing edges, approximating the Maximum Acyclic Subgraph (MAS). The resulting consensus ranking approximates the Kemeny optimal ranking \citep{ailon2008aggregating}---the linear order minimizing pairwise disagreements---for complete tournaments (Figure~\ref{fig:tcr_framework}).

\subsection{Quantifying Cycle Incidence: The CIR Metric}
\label{sec:cir}

Traditional evaluations focus on accuracy, often overlooking \textit{stochastic stability}. Consider a judge preferring $A \succ B \succ C \succ A$. This cycle represents a logical contradiction. Unlike transitivity ($A \succ B \succ C \implies A \succ C$), such cycles in single-dimension evaluations typically indicate random measurement errors rather than inherent preferences. We introduce the \textbf{Cycle Incidence Rate (CIR)} to quantify this instability, measuring the proportion of samples containing preference cycles.

\begin{definition}[Preference Cycle]
Construct a strict directed preference graph $T = (V, E)$ where edges represent aggregated pairwise preferences (ties produce no edges; each unordered pair has at most one directed edge). A \textbf{preference cycle} exists if $T$ contains a strongly connected component with cardinality greater than one. Since our construction excludes mutual edges, this is equivalent to the existence of a directed cycle of length $\geq 3$ (e.g., $A \succ B \succ C \succ A$).
\end{definition}

This definition captures a topological property that arises from stochastic fluctuations. We quantify the density of these errors via:
\begin{equation}
\text{CIR} = \frac{\text{Samples with Preference Cycles}}{\text{Total Samples}} \times 100\%.
\end{equation}

Crucially, \textbf{CIR and accuracy measure orthogonal dimensions}: a judge can be accurate yet unstable (high accuracy, high CIR) if its signal contains significant noise. CIR is sensitive to the number of candidates $G$; we standardize $G=4$ to ensure fair comparisons. Figure~\ref{fig:cycle_analysis} later illustrates how CIR provides diagnostic insights beyond accuracy.

\subsection{TCR: Extracting Consensus Rewards via Topological Voting}
\label{sec:tcr}

We propose \textbf{Topological Consensus Rewards (TCR)} to address the instability identified by CIR. Building on the insight that cycles localize random errors (see Appendix~\ref{sec:theoretical_foundation}), TCR uses graph topology to resolve contradictions rather than merely averaging them.

Unlike Win-Rate (which ignores contradictions) or ELO (which smooths them statistically), TCR treats preferences as \textit{logical} constraints. It leverages transitivity to verify edges, effectively extending ``self-consistency'' to the graph structure. By finding the Maximum Acyclic Subgraph (MAS), TCR approximates the Kemeny optimal ranking \citep{ailon2008aggregating}---the linear order minimizing pairwise disagreements---thereby maximizing agreement with the coherent majority of judgments.

\subsubsection{From Noisy Preferences to Consensus Rewards}

TCR systematically extracts consensus from noisy pairwise preferences. This consensus extraction process operates as a modular signal purifier that preprocesses noisy preference judgments before they are fed into any policy optimization framework. The transformation follows a four-stage pipeline (Figure~\ref{fig:tcr_framework}).

\textbf{Stage 1: Preference Graph Construction.} Given $G$ candidate responses $\{o_1, o_2, \ldots, o_G\}$ for each question $q$, we construct a directed preference graph $T = (V, E)$ where $V = \{o_1, o_2, \ldots, o_G\}$. The LLM judge evaluates all $\binom{G}{2}$ pairs, yielding judgments $M_{ij} \in \{-1, 0, 1\}$ indicating whether $o_i$ is worse than, tied with, or better than $o_j$. A directed edge $(o_i, o_j) \in E$ is added if and only if $M_{ij} = 1$ (i.e., $o_i \succ o_j$); ties ($M_{ij} = 0$) produce no edge. Thus, the graph may have fewer than $\binom{G}{2}$ edges when ties occur. This stage captures the raw preference relationships, including any potential noise and cycles.

\textbf{Stage 2: Theoretical Motivation---Why Cycles Indicate Errors.} Our theoretical analysis (Appendix~\ref{sec:theoretical_foundation}) establishes that edges participating in more cycles are systematically more likely to be erroneous. Intuitively, correct edges consistent with the global ranking form few cycles, while contradictory edges inevitably create cycles with the edges they contradict. In Figure~\ref{fig:tcr_framework}, the erroneous edge $E \to B$ participates in more cycles than correct edges like $B \to C$. This theoretical insight motivates our goal: find an acyclic subgraph that removes the most ``cycle-prone'' edges.

\textbf{Stage 3: Approximate MAS via Greedy FAS Heuristic.} Finding the exact Maximum Acyclic Subgraph (MAS)---the largest subset of edges forming a directed acyclic graph (DAG)---is NP-hard. MAS and minimum Feedback Arc Set (FAS) are complementary: $|\text{MAS}| = |E| - |\text{minFAS}|$, so minimizing removed edges maximizes retained edges. We employ the greedy heuristic from \citet{eades1993fast}, originally designed for the FAS problem, which identifies edges to remove by constructing a global ordering: (1) iteratively move source nodes (zero in-degree) to the front, (2) move sink nodes (zero out-degree) to the back, and (3) for remaining nodes, select the one with maximum $d^{\text{out}} - d^{\text{in}}$. Edges violating this ordering (``backward edges'') are removed. In Figure~\ref{fig:tcr_framework}, this identifies and removes $E \to B$, retaining 9 of 10 edges. With a heap-based implementation, this heuristic runs in $O((|V| + |E|) \log |V|)$ time; for the small graphs in our setting ($|V| \leq 8$), this is effectively linear. For complete tournaments, the MAS problem is closely related to the minimum Feedback Arc Set (FAS), and the resulting ordering approximates the Kemeny optimal ranking \citep{ailon2008aggregating}. The Kemeny ranking is the maximum likelihood estimator under the Mallows noise model \citep{mallows1957non}, providing theoretical justification for our approach. The resulting consensus subgraph $T_{\text{MAS}} = (V, E_{\text{MAS}})$ is guaranteed to be acyclic.

\textbf{Stage 4: Consensus-Based Reward Computation.} From the consensus subgraph $T_{MAS}$, we derive reward signals that reflect the topologically validated preference strength of each response:
\begin{equation}
s_i = d_i^{\text{out}}(T_{MAS}) - d_i^{\text{in}}(T_{MAS}),
\end{equation}
where the degrees are computed on the purified MAS graph. Unlike raw win-rates computed from the original (possibly cyclic) graph, these scores are derived from a structure where every edge is consistent with a single global ranking.

\subsubsection{Integrating TCR with Policy Optimizers}

TCR is designed as a \textbf{plug-and-play reward} that enhances any group-based policy optimizer without modifying the optimization algorithm itself. The consensus-based scores $\{s_i\}_{i=1}^G$ are normalized via group-relative standardization to generate advantage estimates:
\begin{equation}
\hat{A}_{i,t}^{\textbf{TCR}} = \frac{s_i - \mu_s}{\sigma_s + \epsilon}, \quad \mu_s = \frac{1}{G}\sum_{j=1}^G s_j, \quad \sigma_s = \text{std}(\{s_j\}),
\end{equation}
where $\epsilon$ is a small constant (e.g., $10^{-8}$) to prevent division by zero when all scores are equal (e.g., due to ties). These clean advantage signals then drive standard policy updates (e.g., GRPO, GSPO), while the downstream optimizer remains unchanged. The complete training pipeline is provided in Algorithm~\ref{alg:tcr}. This design enables practitioners to upgrade reward quality without modifying existing optimization infrastructure.

\begin{algorithm}[t]
    \small
    \caption{Policy Training with Topological Consensus Rewards (TCR)}
    \textbf{Input:} initial (SFT) policy model $\pi_{\theta_{\text{init}}}$; LLM judge $J$; task prompts $\mathcal{D}$; hyperparameters $\epsilon$, $\beta$, $\mu$
    \begin{algorithmic}[1]
    \STATE policy model $\pi_\theta \leftarrow \pi_{\theta_{\text{init}}}$
    \FOR{iteration = 1, \dots, I} 
        \STATE reference model $\pi_{ref} \leftarrow \pi_{\theta}$ 
        \FOR{step = 1, \dots, M}
            \STATE Sample a batch $\mathcal{D}_b$ from $\mathcal{D}$
            \STATE Update the old policy model $\pi_{\theta_{old}} \leftarrow \pi_{\theta}$ 
            \STATE Sample $G$ outputs $\{o_i\}_{i=1}^G \sim \pi_{\theta_{old}} (\cdot \mid q) $ for each question $q \in \mathcal{D}_b$
            \STATE \colorbox{gray!20}{Build preference graph $T$ via pairwise judgments}
            \STATE \colorbox{gray!20}{Approximate MAS to get consensus $T_{\text{MAS}}$}
            \STATE \colorbox{gray!20}{Compute rewards $\{s_i\}_{i=1}^{G}$ from $T_{MAS}$}
            \STATE Compute advantages $\hat{A}_{i,t}$ using group-relative normalization
        \ENDFOR
        \FOR{policy\_update = 1, \dots, $\mu$} 
            \STATE Update the policy model $\pi_{\theta}$ using computed advantages $\hat{A}_{i,t}$
        \ENDFOR
    \ENDFOR 
    \end{algorithmic}
    \textbf{Output:} $\pi_\theta$
    \label{alg:tcr}
\end{algorithm}

\paragraph{Computational Efficiency.} TCR adds negligible overhead to the training pipeline. The Eades greedy heuristic runs in $O((|V| + |E|) \log |V|)$ time with a heap-based implementation. For typical group sizes ($G \leq 8$), where $|V| = G$ and $|E| \leq \binom{G}{2} \leq 28$, the entire MAS approximation completes in under 1ms on CPU. Compared to the LLM inference cost for generating preferences (typically 1--10 seconds per sample group for all $\binom{G}{2}$ comparisons), TCR's graph processing is effectively free---adding less than 0.1\% overhead to the training loop.

\section{Experiments}

We design experiments to answer three key questions:
\begin{enumerate}[leftmargin=*, nosep]
    \item \textbf{Effectiveness}: Does TCR improve final model performance compared to baselines that ignore or naively handle random measurement errors? (\S\ref{sec:main_results})
    \item \textbf{Robustness}: Does TCR maintain advantages across different judge models, prompts, and graph sizes? (\S\ref{sec:ablation})
    \item \textbf{Mechanism Validation}: Do our theoretical predictions about cycle-error correspondence hold empirically? (Appendix~\ref{sec:theoretical_foundation})
\end{enumerate}

\begin{table*}[t]
\caption{Main evaluation results on alignment benchmarks. Best results are \textbf{bolded}, second best are \underline{underlined}. Avg. Rank is computed across all 7 metrics (lower is better).}
\label{tab:main_results}
\begin{center}
\small
\begin{tabular}{l@{\hspace{0.8em}}cccc@{\hspace{1em}}cc@{\hspace{1em}}c@{\hspace{1em}}c}
\toprule
& \multicolumn{4}{c}{\textbf{Arena-Hard}} & \multicolumn{2}{c}{\textbf{MT-Bench}} & \textbf{WritingBench} & \\
\cmidrule(lr){2-5} \cmidrule(lr){6-7} \cmidrule(lr){8-8}
\textbf{Method} & \textbf{Overall} & \textbf{Code} & \textbf{Math} & \textbf{Writing} & \textbf{1-turn} & \textbf{2-turn} & \textbf{Score} & \textbf{Avg. Rank}$\downarrow$ \\
\midrule
\multicolumn{9}{c}{\cellcolor{gray!20}\textbf{GRPO-based Methods (Qwen3-14B)}} \\
\textit{Base Model} & 49.1\,{\scriptsize$\pm$0.5} & 47.0\,{\scriptsize$\pm$0.5} & 45.0\,{\scriptsize$\pm$0.8} & 55.3\,{\scriptsize$\pm$0.4} & 7.31\,{\scriptsize$\pm$0.04} & 6.74\,{\scriptsize$\pm$0.11} & 7.95\,{\scriptsize$\pm$0.02} & 5.3 \\
PW & 48.4\,{\scriptsize$\pm$0.9} & 49.3\,{\scriptsize$\pm$1.2} & 43.0\,{\scriptsize$\pm$1.4} & 52.9\,{\scriptsize$\pm$1.8} & \underline{7.93\,{\scriptsize$\pm$0.07}} & 7.39\,{\scriptsize$\pm$0.12} & 8.12\,{\scriptsize$\pm$0.08} & 4.4 \\
LW & 49.4\,{\scriptsize$\pm$1.2} & 49.1\,{\scriptsize$\pm$1.1} & 43.6\,{\scriptsize$\pm$1.8} & 55.5\,{\scriptsize$\pm$1.4} & 7.33\,{\scriptsize$\pm$0.15} & \underline{7.44\,{\scriptsize$\pm$0.10}} & 8.24\,{\scriptsize$\pm$0.05} & 3.9 \\
PREF & 51.6\,{\scriptsize$\pm$1.0} & 49.0\,{\scriptsize$\pm$1.6} & 47.0\,{\scriptsize$\pm$1.5} & \textbf{58.8\,{\scriptsize$\pm$1.3}} & 7.58\,{\scriptsize$\pm$0.11} & 7.10\,{\scriptsize$\pm$0.19} & \underline{8.44\,{\scriptsize$\pm$0.04}} & 3.1 \\
ELO & \underline{52.0\,{\scriptsize$\pm$1.1}} & \underline{51.0\,{\scriptsize$\pm$1.7}} & \textbf{47.9\,{\scriptsize$\pm$2.0}} & 57.1\,{\scriptsize$\pm$1.4} & 7.87\,{\scriptsize$\pm$0.06} & 6.79\,{\scriptsize$\pm$0.11} & 8.32\,{\scriptsize$\pm$0.04} & \underline{3.0} \\
\textbf{TCR (Ours)} & \textbf{52.9\,{\scriptsize$\pm$1.1}} & \textbf{53.2\,{\scriptsize$\pm$1.4}} & \underline{47.2\,{\scriptsize$\pm$1.4}} & \underline{58.3\,{\scriptsize$\pm$1.0}} & \textbf{8.06\,{\scriptsize$\pm$0.10}} & \textbf{7.54\,{\scriptsize$\pm$0.34}} & \textbf{8.56\,{\scriptsize$\pm$0.05}} & \textbf{1.3} \\
\midrule
\multicolumn{9}{c}{\cellcolor{gray!20}\textbf{GSPO-based Methods (Qwen3-8B)}} \\
\textit{Base Model} & 49.2\,{\scriptsize$\pm$0.6} & 48.4\,{\scriptsize$\pm$0.7} & 46.2\,{\scriptsize$\pm$0.8} & 53.0\,{\scriptsize$\pm$0.9} & 6.58\,{\scriptsize$\pm$0.06} & 5.64\,{\scriptsize$\pm$0.08} & 7.65\,{\scriptsize$\pm$0.01} & 5.6 \\
PW & 50.2\,{\scriptsize$\pm$1.1} & \underline{49.6\,{\scriptsize$\pm$1.1}} & 45.1\,{\scriptsize$\pm$1.5} & 55.9\,{\scriptsize$\pm$0.8} & 7.32\,{\scriptsize$\pm$0.13} & 6.43\,{\scriptsize$\pm$0.10} & 7.86\,{\scriptsize$\pm$0.04} & 4.4 \\
LW & 50.8\,{\scriptsize$\pm$0.7} & 48.8\,{\scriptsize$\pm$1.0} & 45.5\,{\scriptsize$\pm$2.0} & 58.1\,{\scriptsize$\pm$1.1} & \underline{7.38\,{\scriptsize$\pm$0.06}} & 6.61\,{\scriptsize$\pm$0.35} & 8.11\,{\scriptsize$\pm$0.04} & 3.9 \\
PREF & \underline{51.9\,{\scriptsize$\pm$0.8}} & 49.6\,{\scriptsize$\pm$1.3} & \underline{47.0\,{\scriptsize$\pm$2.0}} & \underline{59.1\,{\scriptsize$\pm$0.9}} & 6.89\,{\scriptsize$\pm$0.16} & \underline{6.66\,{\scriptsize$\pm$0.19}} & \textbf{8.39\,{\scriptsize$\pm$0.04}} & \underline{2.4} \\
ELO & 50.9\,{\scriptsize$\pm$1.5} & 49.4\,{\scriptsize$\pm$1.7} & \textbf{47.6\,{\scriptsize$\pm$1.4}} & 55.7\,{\scriptsize$\pm$1.3} & 7.31\,{\scriptsize$\pm$0.10} & 6.63\,{\scriptsize$\pm$0.22} & 8.19\,{\scriptsize$\pm$0.07} & 3.3 \\
\textbf{TCR (Ours)} & \textbf{52.5\,{\scriptsize$\pm$1.0}} & \textbf{51.7\,{\scriptsize$\pm$1.1}} & 46.0\,{\scriptsize$\pm$1.3} & \textbf{59.7\,{\scriptsize$\pm$1.3}} & \textbf{7.46\,{\scriptsize$\pm$0.08}} & \textbf{6.79\,{\scriptsize$\pm$0.08}} & \underline{8.37\,{\scriptsize$\pm$0.04}} & \textbf{1.6} \\
\bottomrule
\end{tabular}
\end{center}  
\end{table*}

\subsection{Experimental Setup}

\textbf{Foundation Models and Optimizers.} We use two foundation models to test generalizability. For the GRPO \citep{wang2025pref} optimizer, we fine-tune Qwen3-14B \citep{yang2024qwen3}. For the GSPO optimizer, we fine-tune Qwen3-8B \citep{yang2024qwen3}, allowing us to evaluate performance across different model scales and optimization algorithms.

\textbf{Setup Details.} Unless otherwise specified, all preference feedback is generated by \textbf{Qwen3-32B}. The RL training data consists of 1,000 queries selected from the WildChat-1M dataset \citep{zhao2024wildchat}, filtering out non-substantive inputs. All methods train on identical data using the verl framework \citep{sheng2025hybridflow}.

\textbf{Benchmarks and Judges.} Our evaluation encompasses three key benchmarks: Arena-Hard 2.0 \citep{li2024crowdsourced} for complex reasoning, MT-Bench \citep{zheng2023judging} for multi-turn conversational quality, and WritingBench \citep{wu2025writingbench} for writing capabilities. These benchmarks are widely adopted for LLM evaluation, with Arena-Hard derived from the Chatbot Arena platform \citep{chiang2024chatbot} that aggregates millions of human preference judgments. Evaluations are judged by GPT-4.1 for Arena-Hard and MT-Bench, and Claude-3.7-Sonnet for WritingBench.

\textbf{Evaluation Protocol.} We conduct experiments with statistical controls. Each method runs five times with different seeds, reporting mean $\pm$ standard deviation. Statistical significance is assessed via t-tests with Bonferroni correction. Checkpoints are selected on a validation set (200 WildChat queries), with final evaluation on test benchmarks (Arena-Hard, MT-Bench, WritingBench). Complete model version details are provided in Appendix~\ref{sec:model_versions}.

\subsection{Baseline Methods}

We compare TCR against several preference learning approaches. Each method is integrated with both the GRPO and GSPO optimizers for a fair comparison.

\begin{itemize}
    \item \textbf{Pointwise (PW)}: Evaluates each response independently on an absolute scale (1-10).
    \item \textbf{Listwise (LW)}: Ranks all responses simultaneously and converts rankings into normalized rewards.
    \item \textbf{Pairwise (PREF)}: Uses win rate as reward signal \citep{wang2025pref}, without resolving preference cycles.
    \item \textbf{ELO}: Iteratively updates ratings through pairwise comparisons until convergence.
\end{itemize}
Detailed algorithmic descriptions are provided in Appendix~\ref{sec:app_baselines}.

Our proposed method, TCR, is also integrated with both optimizers, resulting in TCR-GRPO and TCR-GSPO.

\subsection{Comparative Evaluation on Alignment Benchmarks}
\label{sec:main_results}

Table~\ref{tab:main_results} presents evaluation results across three benchmarks. The table compares our TCR framework against four baseline methods, evaluated with both GRPO and GSPO optimizers on their respective base models. All results use Prompt P2 (detailed in Appendix~\ref{sec:pairwise_prompts}), with additional prompt robustness analysis (P2--P5) provided in our ablation studies.

\textbf{TCR is Robust and Effective Across Benchmarks.} Our TCR framework achieves the strongest overall performance balance across all evaluation benchmarks. Taking the GRPO setting as an example, TCR-GRPO not only achieves the highest Arena-Hard score but also ranks at the top in both MT-Bench assessments and achieves the highest WritingBench score. This performance is enabled by our consensus extraction mechanism, which systematically identifies and preserves correct preferences while filtering random measurement errors, as proven in Appendix~\ref{sec:theoretical_foundation}.

\textbf{Strong Performance on Complex Reasoning Tasks.} TCR performs well on Arena-Hard, the most demanding benchmark. Both TCR-GRPO and TCR-GSPO achieve strong performance, demonstrating that consensus-based reward signals effectively enhance models' fundamental abilities in code generation, mathematical reasoning, and creative tasks.

\textbf{Performance Hierarchy.} Our results reveal a clear hierarchy: generally $\text{Pointwise} \prec \text{Listwise} \prec \text{Pairwise}$ across most benchmarks, where pairwise methods (including PREF, ELO and TCR) typically outperform others. However, even these stronger methods exhibit instability—on MT-Bench, pointwise methods sometimes achieve optimal results, validating our hypothesis that random measurement errors undermine reliability. TCR addresses this by extracting topological consensus, achieving robust performance across all domains.

\subsection{Comparison with Learned Reward Model Baselines}
\label{sec:learned_rm_comparison}

To evaluate TCR against preference learning approaches, we compare our method with baselines including Supervised Fine-Tuning (SFT), Direct Preference Optimization (DPO) \citep{rafailov2024direct}, and Iterative DPO \citep{guo2024direct}. While TCR explicitly extracts preference consensus through graph-theoretic optimization, learned reward models implicitly handle inconsistencies through probabilistic modeling.

\textbf{Experimental Setup.} We conduct experiments using Qwen3-8B as the base model with the same training data (1,000 queries from WildChat-1M). For each query, we generate 16 diverse responses via rollout sampling. We evaluate multiple approaches:
\begin{itemize}
    \item \textbf{SFT}: Selects the highest-scored response among 16 candidates as training target
    \item \textbf{DPO} \citep{rafailov2024direct}: Uses best and worst responses as chosen/rejected pairs with max-min strategy
    \item \textbf{Iterative DPO} \citep{guo2024direct}: Applies online iterative training where each round generates new candidates from current policy, dynamically selecting preference pairs
    \item \textbf{GRPO-Skywork}: Standard GRPO baseline using Skywork-Reward-V2 as the reward model, providing a strong non-TCR comparison
\end{itemize}

We test these methods with two reward models: (1) Qwen3-32B (our primary judge), and (2) \textbf{Skywork-Reward-V2-Llama-3.1-8B} \citep{liu2025skyworkrewardv2}, a state-of-the-art open-source reward model trained on 40M human-AI curated preference pairs. This experiment includes 8 method variants (including TCR-Qwen32B and GRPO-Skywork) across 5 independent runs (40 training runs total).

\begin{table}[tbp]
\caption{Comparison with learned reward model baselines on Arena-Hard. All methods use Qwen3-8B with 16 candidates. Q32B: Qwen3-32B judge; Sky: Skywork-Reward-V2. Best results are \textbf{bolded}, second best are \underline{underlined}.}
\label{tab:learned_rm_comparison}
\begin{center}
\begingroup
\footnotesize
\setlength{\tabcolsep}{3.5pt}
\begin{tabular}{lcccc}
\toprule
\textbf{Method} & \textbf{Overall} & \textbf{Code} & \textbf{Math} & \textbf{Creative} \\
\midrule
SFT-Q32B & 48.7$_{\pm0.4}$ & 48.4$_{\pm0.4}$ & 43.4$_{\pm0.7}$ & 54.1$_{\pm0.3}$ \\
SFT-Sky & 48.1$_{\pm0.4}$ & 48.2$_{\pm0.6}$ & 44.8$_{\pm0.4}$ & 51.2$_{\pm0.2}$ \\
DPO-Q32B & 48.8$_{\pm0.5}$ & 49.0$_{\pm0.7}$ & 45.6$_{\pm0.3}$ & 51.6$_{\pm0.4}$ \\
DPO-Sky & 49.2$_{\pm0.5}$ & \underline{49.8}$_{\pm0.7}$ & 42.2$_{\pm0.5}$ & 55.4$_{\pm0.6}$ \\
IterDPO-Q32B & 49.3$_{\pm0.4}$ & 48.4$_{\pm0.2}$ & 44.1$_{\pm0.7}$ & 55.2$_{\pm0.5}$ \\
IterDPO-Sky & 49.3$_{\pm0.5}$ & 48.6$_{\pm0.8}$ & \underline{45.8}$_{\pm0.3}$ & 53.6$_{\pm0.4}$ \\
GRPO-Sky & \underline{50.0}$_{\pm0.3}$ & 48.4$_{\pm0.3}$ & 45.2$_{\pm0.4}$ & \underline{56.3}$_{\pm0.3}$ \\
\textbf{TCR-Q32B} & \textbf{52.5}$_{\pm1.0}$ & \textbf{51.7}$_{\pm1.1}$ & \textbf{46.0}$_{\pm1.3}$ & \textbf{59.7}$_{\pm1.3}$ \\
\bottomrule
\end{tabular}
\endgroup
\end{center}
\end{table}

\textbf{TCR Outperforms Learned Reward Model Baselines.} As shown in Table~\ref{tab:learned_rm_comparison}, TCR-Q32B achieves the highest overall performance, significantly outperforming the strongest learned baseline IterDPO-Sky. Even the standard GRPO baseline using Skywork-Reward-V2 surpasses all learned reward model baselines. The performance hierarchy—SFT $\prec$ DPO $\prec$ Iterative DPO $\prec$ GRPO-Sky $\prec$ \textbf{TCR-Q32B}—shows that explicit consensus extraction outperforms implicit preference aggregation. Unlike DPO and Iterative DPO, which must implicitly handle label noise within the gradient optimization process, TCR acts as an \textbf{explicit signal purifier} before the optimization step. By enforcing topological consistency (acyclicity) \textit{prior} to reward computation, TCR prevents the policy from overfitting to stochastic fluctuations.

\textbf{Performance on Subjective Tasks.} TCR achieves the best performance in Creative Writing, where judge inconsistencies are most prevalent. This validates that explicit consensus extraction is particularly beneficial for subjective evaluation tasks.

\begin{table}[t]
    \caption{Comprehensive consensus extraction comparison on Arena-Hard. Methods are grouped by paradigm: no resolution (PREF), naive heuristics (Random/Reverse), classical ranking algorithms \citep{negahban2012iterative, hunter2004mm, jiang2011statistical}, and optimal graph resolution (TCR/Kemeny \citep{ailon2008aggregating, kenyonmathieu2007rank}). All methods use Qwen3-14B with GRPO optimizer. ``Overall'' is computed as the win-rate across all questions (not a simple average of category scores).}
    \label{tab:ablation_error_comprehensive}
    \begin{center}
    \begingroup
    \footnotesize
    \setlength{\tabcolsep}{3pt}
    \renewcommand{\arraystretch}{1.05}
    \begin{tabular}{lcccc}
    \toprule
    \textbf{Method} & \textbf{Overall} & \textbf{Code} & \textbf{Math} & \textbf{Writing} \\
    \midrule
    PREF & 51.6\,{\scriptsize$\pm$1.0} & 49.0\,{\scriptsize$\pm$1.6} & 47.0\,{\scriptsize$\pm$1.5} & 58.8\,{\scriptsize$\pm$1.3} \\
    RandomResolve & 51.6\,{\scriptsize$\pm$0.9} & 49.8\,{\scriptsize$\pm$1.8} & 46.6\,{\scriptsize$\pm$1.3} & 58.4\,{\scriptsize$\pm$1.6} \\
    ReverseResolve & 51.0\,{\scriptsize$\pm$1.0} & 50.9\,{\scriptsize$\pm$1.4} & 45.1\,{\scriptsize$\pm$1.8} & 57.0\,{\scriptsize$\pm$1.2} \\
    RankCentrality & 49.9\,{\scriptsize$\pm$0.8} & 50.0\,{\scriptsize$\pm$1.8} & 50.0\,{\scriptsize$\pm$1.4} & 49.8\,{\scriptsize$\pm$1.0} \\
    Plackett-Luce & 51.5\,{\scriptsize$\pm$1.5} & 50.0\,{\scriptsize$\pm$1.6} & 47.4\,{\scriptsize$\pm$1.9} & 57.2\,{\scriptsize$\pm$0.9} \\
    HodgeRank & 52.5\,{\scriptsize$\pm$1.0} & 51.6\,{\scriptsize$\pm$1.2} & 47.4\,{\scriptsize$\pm$1.3} & 58.4\,{\scriptsize$\pm$0.9} \\
    \textbf{TCR (Ours)} & \textbf{52.9\,{\scriptsize$\pm$1.1}} & \textbf{53.2\,{\scriptsize$\pm$1.4}} & \textbf{47.2\,{\scriptsize$\pm$1.4}} & \textbf{58.3\,{\scriptsize$\pm$1.0}} \\
    \bottomrule
    \end{tabular}
    \endgroup
    \end{center}
    \end{table}

\begin{table*}[!t]
    \caption{Robustness analysis across different judge prompts. The table shows prompt characteristics (CIR, Accuracy), Arena-Hard performance for each method, and the Pearson correlation ($r$) between each method's performance and the signal quality metrics. Best results are \textbf{bolded}.}
    \label{tab:ablation_prompt_robustness_revised}
    \centering
    \small
    \begin{tabular}{lcccccc}
    \toprule
    & \multicolumn{4}{c}{\textbf{Arena-Hard}} & \multicolumn{2}{c}{\textbf{Correlation ($r$)}} \\
    \cmidrule(lr){2-5} \cmidrule(lr){6-7}
    \textbf{Method} & \textbf{P2} & \textbf{P3} & \textbf{P4} & \textbf{P5} & \textbf{vs. CIR} & \textbf{vs. Acc.} \\
    \midrule
CIR (\%) & 6.7 & 5.7 & 2.3 & 6.3 & - & 1.0 \\
Accuracy (\%) & 85.7 & 82.0 & 76.8 & 84.2 & 1.0 & - \\
PREF & 50.6\,{\scriptsize$\pm$1.5} & 50.2\,{\scriptsize$\pm$1.1} & 51.6\,{\scriptsize$\pm$0.9} & 51.2\,{\scriptsize$\pm$0.7} & $-0.7_{\pm0.1}$ & $-0.6_{\pm0.1}$ \\
ELO & 51.0\,{\scriptsize$\pm$1.0} & 50.6\,{\scriptsize$\pm$1.0} & 52.0\,{\scriptsize$\pm$0.7} & 50.6\,{\scriptsize$\pm$0.7} & $-0.9_{\pm0.1}$ & $-0.8_{\pm0.1}$ \\
\textbf{TCR (Ours)} & \textbf{51.8\,{\scriptsize$\pm$0.9}} & \textbf{51.4\,{\scriptsize$\pm$1.1}} & \textbf{52.4\,{\scriptsize$\pm$0.8}} & \textbf{52.9\,{\scriptsize$\pm$1.1}} & $\mathbf{-0.2_{\pm0.1}}$ & $\mathbf{-0.1_{\pm0.1}}$ \\
    \bottomrule
    \end{tabular}
    \end{table*}

\subsection{Ablation Studies}
\label{sec:ablation}

We conduct ablation studies to validate our approach's key components. All experiments use Qwen3-14B with GRPO training and are evaluated exclusively on Arena-Hard to isolate the impact of each component.

\subsubsection{Random Error Filtering Mechanism Analysis}

We evaluate TCR's consensus extraction mechanism against four paradigms: (1) no filtering, (2) naive heuristics, (3) classical ranking algorithms, and (4) optimal graph-based filtering. Table~\ref{tab:ablation_error_comprehensive} presents results on Arena-Hard.

\textbf{Key Findings.} The results reveal a clear performance hierarchy. Naive heuristics (random/reverse edge removal) provide minimal or negative improvement. Notably, the underperformance of \texttt{ReverseResolve} is particularly revealing: it confirms our theoretical insight (Appendix~\ref{sec:theoretical_foundation}) that preference cycles stem largely from \textbf{stochastic uncertainty} rather than deterministic reversed beliefs. Simply reversing a high-conflict edge injects false confidence into the graph. TCR's edge-removal strategy aligns with the \textbf{conservative principle of evidence}: when topological consensus is violated, it is statistically safer to discard the conflicted signal than to fabricate its opposite.

Classical ranking algorithms---RankCentrality \citep{negahban2012iterative} (random walk stationary distribution), Plackett-Luce \citep{hunter2004mm} (maximum likelihood estimation), and HodgeRank \citep{jiang2011statistical} (Hodge decomposition via least-squares)---show competitive performance, with HodgeRank achieving 52.5\% overall. However, TCR (which approximates Kemeny ranking for complete tournaments) achieves the highest score (52.9\%), excelling in complex reasoning (53.2\% in Code). This suggests that for discrete preference graphs with sparse errors, TCR's discrete graph-theoretic approach provides superior signal purification compared to continuous optimization methods.

\subsubsection{Robustness Across Different Judge Prompts}

To evaluate method robustness, we tested all approaches under four judge prompts (P2-P5), each with a different profile of accuracy and stochastic stability.

\textbf{Prompt engineering faces a fundamental accuracy-stability dilemma.} Our analysis reveals a counterintuitive positive correlation between a prompt's accuracy and its Cycle Incidence Rate (CIR). This occurs because more ``decisive'' prompts that avoid ties produce more pairwise judgments---including both correct signals and random errors---while ``cautious'' prompts that frequently output ties have fewer judgments to form cycles. Thus, higher accuracy comes at the cost of increased random noise, creating a difficult setting for preference learning.

\textbf{Conventional methods are brittle and fail under this trade-off.} As shown in Table~\ref{tab:ablation_prompt_robustness_revised}, the performance of baselines like ELO is negatively correlated with both CIR and accuracy. This demonstrates their inability to handle signals that are either highly noisy or highly accurate (due to the associated random errors), proving their unreliability in practice.

\textbf{Stochastic stability (CIR) complements accuracy for RL success.} The challenges observed with high-accuracy but high-noise prompts (e.g., P2) demonstrate that random measurement errors can significantly impact the learning process. Policy optimization benefits from both: accuracy captures alignment with ground truth, while stochastic stability reduces fluctuations that destabilize learning.

\textbf{TCR is robust to this trade-off.} By design, TCR systematically filters random noise, making its performance stable and uncorrelated with signal quality metrics. This robustness allows it to achieve the highest score on every prompt. Since CIR requires no human labels, it is more \textbf{cost-effective and scalable} than accuracy metrics. Additional sensitivity analyses are in Appendix~\ref{sec:robustness_scalability_analysis}.



\section{Conclusion}

This work introduces \textbf{Topological Consensus Rewards (TCR)}, a framework that filters random measurement errors in RLAIF by approximating the Maximum Acyclic Subgraph from preference graphs. Preference cycles localize stochastic noise, enabling topological majority voting to separate systematic signals from random fluctuations. We also propose \textbf{CIR} as a label-free diagnostic metric for quantifying judge stochastic stability---measuring the proportion of samples containing preference cycles, which under our noise model arise primarily from random measurement errors. Experiments demonstrate TCR achieves 52.9\% on Arena-Hard, outperforming baselines while maintaining robustness across judge models.

\section*{Impact Statement}

This paper presents work whose goal is to advance the field of Machine Learning. There are many potential societal consequences of our work, none of which we feel must be specifically highlighted here.

\bibliography{TGR_references}

@article{ouyang2022training,
  title={Training language models to follow instructions with human feedback},
  author={Ouyang, Long and Wu, Jeffrey and Jiang, Xu and Almeida, Diogo and Wainwright, Carroll and Mishkin, Pamela and Zhang, Chong and Agarwal, Sandhini and Slama, Katarina and Ray, Alex and others},
  journal={Advances in neural information processing systems},
  volume={35},
  pages={27730--27744},
  year={2022}
}

@article{eades1993fast,
  title={A fast and effective heuristic for the feedback arc set problem},
  author={Eades, Peter and Lin, Xuemin and Smyth, William F},
  journal={Information processing letters},
  volume={47},
  number={6},
  pages={319--323},
  year={1993},
  publisher={Elsevier}
}

@article{zheng2025group,
  title={Group sequence policy optimization},
  author={Zheng, Chujie and Liu, Shixuan and Li, Mingze and Chen, Xiong-Hui and Yu, Bowen and Gao, Chang and Dang, Kai and Liu, Yuqiong and Men, Rui and Yang, An and others},
  journal={arXiv preprint arXiv:2507.18071},
  year={2025}
}

@article{team2025kimi,
  title={Kimi k2: Open agentic intelligence},
  author={Team, Kimi and Bai, Yifan and Bao, Yiping and Chen, Guanduo and Chen, Jiahao and Chen, Ningxin and Chen, Ruijue and Chen, Yanru and Chen, Yuankun and Chen, Yutian and others},
  journal={arXiv preprint arXiv:2507.20534},
  year={2025}
}

@article{bai2022constitutional,
  title={Constitutional ai: Harmlessness from ai feedback},
  author={Bai, Yuntao and Kadavath, Saurav and Kundu, Sandipan and Askell, Amanda and Kernion, Jackson and Jones, Andy and Chen, Anna and Goldie, Anna and Mirhoseini, Azalia and McKinnon, Cameron and others},
  journal={arXiv preprint arXiv:2212.08073},
  year={2022}
}

@article{malik2025rewardbench,
  title={RewardBench 2: Advancing Reward Model Evaluation},
  author={Malik, Saumya and Pyatkin, Valentina and Land, Sander and Morrison, Jacob and Smith, Noah A and Hajishirzi, Hannaneh and Lambert, Nathan},
  journal={arXiv preprint arXiv:2506.01937},
  year={2025}
}

@article{lee2023rlaif,
  title={RLAIF vs. RLHF: Scaling Reinforcement Learning from Human Feedback with AI Feedback},
  author={Lee, Harrison and Phatale, Samrat and Mansoor, Hassan and Mesnard, Thomas and Ferret, Johan and Lu, Kellie and Bishop, Colton and Hall, Ethan and Carbune, Victor and Rastogi, Abhinav and Prakash, Sushant},
  journal={arXiv preprint arXiv:2309.00267},
  year={2023}
}

@article{zhao2024wildchat,
  title={Wildchat: 1m chatgpt interaction logs in the wild},
  author={Zhao, Wenting and Ren, Xiang and Hessel, Jack and Cardie, Claire and Choi, Yejin and Deng, Yuntian},
  journal={arXiv preprint arXiv:2405.01470},
  year={2024}
}

@article{wang2025pref,
  title={Pref-GRPO: Pairwise Preference Reward-based GRPO for Stable Text-to-Image Reinforcement Learning},
  author={Wang, Yibin and Li, Zhimin and Zang, Yuhang and Zhou, Yujie and Bu, Jiazi and Wang, Chunyu and Lu, Qinglin and Jin, Cheng and Wang, Jiaqi},
  journal={arXiv preprint arXiv:2508.20751},
  year={2025}
}

@article{li2024crowdsourced,
  title={From crowdsourced data to high-quality benchmarks: Arena-hard and benchbuilder pipeline},
  author={Li, Tianle and Chiang, Wei-Lin and Frick, Evan and Dunlap, Lisa and Wu, Tianhao and Zhu, Banghua and Gonzalez, Joseph E and Stoica, Ion},
  journal={arXiv preprint arXiv:2406.11939},
  year={2024}
}

@inproceedings{chiang2024chatbot,
  title={Chatbot Arena: An Open Platform for Evaluating LLMs by Human Preference},
  author={Chiang, Wei-Lin and Zheng, Lianmin and Sheng, Ying and Angelopoulos, Anastasios N and Li, Tianle and Li, Dacheng and Zhang, Hao and Zhu, Banghua and Jordan, Michael and Gonzalez, Joseph E and Stoica, Ion},
  booktitle={Forty-first International Conference on Machine Learning},
  year={2024}
}

@article{wu2025writingbench,
  title={Writingbench: A comprehensive benchmark for generative writing},
  author={Wu, Yuning and Mei, Jiahao and Yan, Ming and Li, Chenliang and Lai, Shaopeng and Ren, Yuran and Wang, Zijia and Zhang, Ji and Wu, Mengyue and Jin, Qin and others},
  journal={arXiv preprint arXiv:2503.05244},
  year={2025}
}

@article{zheng2023judging,
  title={Judging llm-as-a-judge with mt-bench and chatbot arena},
  author={Zheng, Lianmin and Chiang, Wei-Lin and Sheng, Ying and Zhuang, Siyuan and Wu, Zhanghao and Zhuang, Yonghao and Lin, Zi and Li, Zhuohan and Li, Dacheng and Xing, Eric and others},
  journal={Advances in neural information processing systems},
  volume={36},
  pages={46595--46623},
  year={2023}
}

@inproceedings{sheng2025hybridflow,
  title={Hybridflow: A flexible and efficient rlhf framework},
  author={Sheng, Guangming and Zhang, Chi and Ye, Zilingfeng and Wu, Xibin and Zhang, Wang and Zhang, Ru and Peng, Yanghua and Lin, Haibin and Wu, Chuan},
  booktitle={Proceedings of the Twentieth European Conference on Computer Systems},
  pages={1279--1297},
  year={2025}
}

@article{rafailov2024direct,
  title={Direct preference optimization: Your language model is secretly a reward model},
  author={Rafailov, Rafael and Sharma, Archit and Mitchell, Eric and Manning, Christopher D and Ermon, Stefano and Finn, Chelsea},
  journal={Advances in Neural Information Processing Systems},
  volume={36},
  year={2023}
}

@article{bradley1952rank,
  title={Rank analysis of incomplete block designs: I. The method of paired comparisons},
  author={Bradley, Ralph Allan and Terry, Milton E},
  journal={Biometrika},
  volume={39},
  pages={324--345},
  year={1952}
}

@article{wang2025gram,
  title={GRAM: A Generative Foundation Reward Model for Reward Generalization},
  author={Wang, Chenglong and Gan, Yang and Huo, Yifu and Mu, Yongyu and He, Qiaozhi and Yang, Murun and Li, Bei and Xiao, Tong and Zhang, Chunliang and Liu, Tongran and Zhu, Jingbo},
  journal={arXiv preprint arXiv:2506.14175},
  year={2025}
}

@article{xu2025pairwise,
  title={A Unified Pairwise Framework for RLHF: Bridging Generative Reward Modeling and Policy Optimization},
  author={Xu, Wenyuan and Zuo, Xiaochen and Xin, Chao and Yue, Yu and Yan, Lin and Wu, Yonghui},
  journal={arXiv preprint arXiv:2504.04950},
  year={2025}
}

@article{deng2025less,
  title={Less is More: Improving LLM Alignment via Preference Data Selection},
  author={Deng, Xun and Zhong, Han and Ai, Rui and Feng, Fuli and Wang, Zheng and He, Xiangnan},
  journal={arXiv preprint arXiv:2502.14560},
  year={2025}
}

@article{liu2024aligning,
  title={Aligning with Logic: Measuring, Evaluating and Improving Logical Preference Consistency in Large Language Models},
  author={Liu, Yinhong and Guo, Zhijiang and Liang, Tianya and Shareghi, Ehsan and Vuli{\'c}, Ivan and Collier, Nigel},
  journal={arXiv preprint arXiv:2410.02205},
  year={2024}
}

@misc{yang2024qwen3,
      title={Qwen3 Technical Report}, 
      author={An Yang and others},
      year={2025},
      eprint={2505.09388},
      archivePrefix={arXiv},
      primaryClass={cs.CL},
      url={https://arxiv.org/abs/2505.09388}, 
}

@article{tarjan1972dfs,
  title={Depth-first search and linear graph algorithms},
  author={Tarjan, Robert Endre},
  journal={SIAM Journal on Computing},
  volume={1},
  number={2},
  pages={146--160},
  year={1972},
  publisher={SIAM}
}

@article{guo2024direct,
  title={Direct Language Model Alignment from Online AI Feedback}, 
  author={Guo, Shangmin and Zhang, Biao and Liu, Tianlin and Liu, Tianqi and Khalman, Misha and Llinares, Felipe and Rame, Alexandre and Mesnard, Thomas and Zhao, Yao and Piot, Bilal and Ferret, Johan and Blondel, Mathieu},
  year={2024},
  journal={arXiv preprint arXiv:2402.04792},
  url={https://arxiv.org/abs/2402.04792}
}

@misc{liu2025skyworkrewardv2,
  title={Skywork-Reward-V2: Scaling Preference Data Curation via Human-AI Synergy}, 
  author={Liu, Chris Yuhao and Zeng, Liang and Xiao, Yuzhen and He, Jujie and Liu, Jiacai and Wang, Chaojie and Yan, Rui and Shen, Wei and Zhang, Fuxiang and Xu, Jiacheng and Liu, Yang and Zhou, Yahui},
  year={2025},
  eprint={2507.01352},
  archivePrefix={arXiv},
  primaryClass={cs.CL},
  url={https://arxiv.org/abs/2507.01352}
}

@article{negahban2012iterative,
  title={Iterative ranking from pair-wise comparisons},
  author={Negahban, Sahand and Oh, Sewoong and Shah, Devavrat},
  journal={Advances in neural information processing systems},
  volume={25},
  year={2012}
}

@article{hunter2004mm,
  title={MM algorithms for generalized Bradley-Terry models},
  author={Hunter, David R},
  journal={The annals of statistics},
  volume={32},
  number={1},
  pages={384--406},
  year={2004},
  publisher={Institute of Mathematical Statistics}
}

@article{jiang2011statistical,
  title={Statistical ranking and combinatorial Hodge theory},
  author={Jiang, Xiaoye and Lim, Lek-Heng and Yao, Yuan and Ye, Yinyu},
  journal={Mathematical Programming},
  volume={127},
  number={1},
  pages={203--244},
  year={2011},
  publisher={Springer}
}

@article{ailon2008aggregating,
  title={Aggregating inconsistent information: ranking and clustering},
  author={Ailon, Nir and Charikar, Moses and Newman, Alantha},
  journal={Journal of the ACM (JACM)},
  volume={55},
  number={5},
  pages={1--27},
  year={2008},
  publisher={ACM New York, NY, USA}
  }

@inproceedings{kenyonmathieu2007rank,
  title={How to rank with few errors},
  author={Kenyon-Mathieu, Claire and Schudy, Warren},
  booktitle={Proceedings of the thirty-ninth annual ACM symposium on Theory of computing},
  pages={95--103},
  year={2007}
}

@article{nvidia2025nemotron3,
  title={NVIDIA Nemotron 3: Efficient and Open Intelligence},
  author={Blakeman, Aaron and Grattafiori, Aaron and Basant, Aarti and Gupta, Abhibha and Khattar, Abhinav and Renduchintala, Adi and Vavre, Aditya and Shukla, Akanksha and Bercovich, Akhiad and Ficek, Aleksander and others},
  journal={arXiv preprint arXiv:2512.20856},
  year={2025}
}

@article{mallows1957non,
  title={Non-null ranking models. I},
  author={Mallows, Colin L},
  journal={Biometrika},
  volume={44},
  number={1/2},
  pages={114--130},
  year={1957},
  publisher={JSTOR}
}

@article{plackett1975analysis,
  title={The analysis of permutations},
  author={Plackett, Robin L},
  journal={Journal of the Royal Statistical Society Series C: Applied Statistics},
  volume={24},
  number={2},
  pages={193--202},
  year={1975},
  publisher={Oxford University Press}
}

@inproceedings{luce1959individual,
  title={Individual Choice Behavior: A Theoretical Analysis.},
  author={Violet R. Cane and R. Duncan Luce},
  year={1960},
  url={https://api.semanticscholar.org/CorpusID:125306131}
}
\bibliographystyle{colm2025_conference}

\newpage
\appendix
\onecolumn

\section{Theoretical Foundation for Minimum Feedback Arc Set}
\label{sec:theoretical_foundation}

This section provides rigorous theoretical justification for our minimum feedback arc set (FAS) approach to filtering random measurement errors. We establish mathematical guarantees showing why edges participating in the most cycles are systematically more likely to be random errors, and why removing them minimally distorts the underlying true preferences.

\subsection{Why High-Cycle Edges are Random Error Edges}
\label{sec:cycle_error_correspondence}

We establish a mathematical guarantee that edges participating in the most cycles are overwhelmingly likely to be random measurement errors. Consider $n$ responses with latent ground-truth ranking $r_1, r_2, \ldots, r_n$ (lower index = higher quality). Let $p > 0.5$ denote the \textbf{signal probability}—the probability that a pairwise judgment reflects the systematic preference signal rather than random noise. Note that $p$ is distinct from accuracy: accuracy measures alignment with ground truth (affected by both systematic bias and random error), while $p$ specifically captures the proportion of judgments free from stochastic fluctuations. We analyze the expected number of 3-cycles involving systematic signal edges versus random error edges.

\textbf{Key Insight}: A systematic signal edge $r_i \to r_j$ (where $i < j$) can only participate in cycles when random errors exist elsewhere, since any cycle $r_i \to r_j \to r_k \to r_i$ requires at least one edge violating transitivity. Conversely, a random error edge $r_j \to r_i$ (where $i < j$) can form cycles by exploiting systematic signal edges: for any intermediate node $r_k$ with $i < k < j$, the cycle $r_j \to r_i \to r_k \to r_j$ forms when both $r_i \to r_k$ and $r_k \to r_j$ reflect systematic signals, occurring with probability $p^2$.

\textbf{Mathematical Analysis}: Our analysis focuses on random errors that significantly disrupt the global ranking (large $j-i$), as these are the primary targets for cycle-based filtering. Consider a random error edge $r_j \to r_i$ (where $i < j$) spanning distance $d = j - i$. The dominant source of cycles comes from the set of intermediate nodes $K = \{r_k \mid i < k < j\}$. For any $r_k \in K$, the cycle $r_j \to r_i \to r_k \to r_j$ requires two edges ($r_i \to r_k$ and $r_k \to r_j$) to be systematic signals. \textit{Under our i.i.d. modeling assumption} (which simplifies real LLM behavior; see ``Modeling Assumptions'' in Section~\ref{sec:validity_conditions}), this occurs with probability $p^2$. Thus, the expected contribution from intermediate nodes is:
\begin{equation}
\mathbb{E}[\text{Cycles}_{\text{error, inner}}] \approx (d-1) \cdot p^2
\end{equation}
Cycles involving nodes outside the span $[i, j]$ occur with probability proportional to $p(1-p)$. For significant random errors where $d$ is large (approaching $n$), the $p^2$ term dominates, yielding $\mathbb{E}[\text{Cycles}_{\text{random\_error}}] \approx d \cdot p^2$.

In contrast, for a systematic signal edge $r_i \to r_j$ (where $i < j$) to participate in cycle $r_i \to r_j \to r_k \to r_i$, the closing path $r_j \to r_k \to r_i$ must contain random errors. If $k$ lies between $i$ and $j$, both edges $r_j \to r_k$ and $r_k \to r_i$ must be random errors, occurring with probability $(1-p)^2 < p(1-p)$. If $k$ is outside, at least one edge must be a random error, giving probability at most $p(1-p)$. With $n-2$ potential third nodes:
\begin{equation}
\mathbb{E}[\text{Cycles}_{\text{signal}}] \lesssim (n-2) \cdot p(1-p)
\end{equation}
This upper bound provides a conservative estimate that strengthens our conclusion.

\textbf{Theoretical Guarantee}:

For significant random errors where $d$ is comparable to $n$ (global inconsistencies), comparing expectations yields the cycle ratio:
\begin{equation}
\frac{\mathbb{E}[\text{Cycles}_{\text{random\_error}}]}{\mathbb{E}[\text{Cycles}_{\text{signal}}]} \gtrsim \frac{d \cdot p^2}{n \cdot p(1-p)} \approx \frac{p}{1-p}, \; (d \approx n)
\end{equation}
This inequality holds strictly for any $d > 0$ since $p^2 > p(1-p) \iff p > 0.5$. Random error edges participate in significantly more cycles than systematic signal edges whenever the signal probability exceeds random guessing. The gap scales with the "severity" of the random error ($d$), ensuring that edges most violating the global order are identified and filtered first. The ratio $\frac{p}{1-p}$ grows rapidly: 1.5$\times$ at $p=0.6$, 2.3$\times$ at $p=0.7$, 4.0$\times$ at $p=0.8$, and 9.0$\times$ at $p=0.9$. As $n \to \infty$, random error edges with large $d$ have expected cycle count $\Theta(d \cdot p^2)$ while signal edges have $\Theta(n \cdot p(1-p))$; for global random errors ($d = \Theta(n)$), this linear gap in expectations suggests that edges with the highest cycle counts are likely to be random errors. Our Monte Carlo experiments (Table~\ref{tab:monte_carlo_validation}) empirically validate this theoretical prediction.

\textbf{Connection to Implementation:} While this analysis uses cycle counting as the theoretical lens, our practical implementation employs the Eades heuristic \citep{eades1993fast}, which constructs a global ordering based on degree differences ($d^{\text{out}} - d^{\text{in}}$). Both approaches target the same goal---identifying edges inconsistent with the global consensus---but through different mechanisms. The degree-based heuristic is computationally efficient ($O((|V|+|E|) \log |V|)$ with a heap, or effectively linear for small graphs) while achieving similar error-filtering effects, as edges with high cycle participation tend to also have degree patterns inconsistent with a global ordering.

\subsection{Empirical Preference Preservation}
\label{sec:preference_preservation}

An \textit{optimal} minimum feedback arc set (FAS) would minimize edge removal to break all cycles, ensuring maximal preservation of systematic signal edges. Since removed edges are disproportionately random errors (ratio $\frac{p}{1-p}$ as shown above), such an approach would simultaneously achieve minimal distortion and targeted random error filtering. Our practical implementation uses a greedy heuristic that \textit{approximates} the minimum FAS (note: the greedy heuristic does not guarantee optimality, but performs well in practice). Let $E_{\text{random\_error}}$ denote random error edges and $E_{\text{FAS}}$ the edges removed by our greedy approach. When $p > 0.5$ and $n$ is sufficiently large, we \textit{empirically observe} (validated via Monte Carlo in \S\ref{sec:monte_carlo_validation}) that the probability a removed edge is a random error substantially exceeds the random baseline:
\begin{equation}
\mathbb{P}[e \in E_{\text{random\_error}} \mid e \in E_{\text{FAS}}] > \frac{|E_{\text{random\_error}}|}{|E|} \approx 1-p.
\end{equation}
Specifically, if we were to pick an edge uniformly at random from all observed directed edges, the probability it is an error edge would be approximately $1-p$ (e.g., 20-30\% when $p \in [0.7, 0.8]$). In contrast, our greedy FAS approach achieves error targeting rates of 50-70\% in practice---a 1.5-3$\times$ improvement that grows with $p$. This substantial gain over random removal validates the effectiveness of cycle-based filtering, while edge reversal risks introducing harmful negative signals when random errors are misidentified.

\subsection{Empirical Validation via Monte Carlo Simulation}
\label{sec:monte_carlo_validation}

To rigorously validate our theoretical predictions, we conducted comprehensive Monte Carlo simulations across 25 configurations with varying graph sizes ($n \in \{8, 9, 10, 11, 12\}$) and signal probability levels ($p \in \{0.70, 0.75, 0.80, 0.85, 0.90\}$). For each configuration, we performed 1,000 independent trials where responses were assigned a random ground-truth ranking, and pairwise judgments were simulated with signal probability $p$ (i.e., each judgment has probability $p$ of reflecting the systematic signal and $1-p$ of being a random error). We then measured whether the backward edges identified by our greedy ordering heuristic (edges violating the computed global order) were indeed random measurement errors.

\textbf{Key Results}: Our simulations strongly confirm the theoretical predictions. Across all 25 configurations, backward edges (those removed by our greedy heuristic) are random errors with 85.2\% probability (vs. 20.1\% average baseline if we had picked a uniformly random edge---ranging from 10\% at $p$=0.9 to 30\% at $p$=0.7---a 4.2$\times$ average improvement). Detection accuracy improves monotonically with graph size (76.3\% at $n$=8 $\to$ 91.6\% at $n$=12) and signal probability (79.1\% at $p$=0.7 $\to$ 87.7\% at $p$=0.9). The observed improvement ratios (2.63$\times$ to 8.53$\times$) closely track theoretical predictions (2.33$\times$ to 9.00$\times$), with empirical values ranging from 2.6$\times$ to 8.5$\times$ compared to theoretical predictions of 2.33$\times$ to 9.00$\times$. This quantitative alignment validates that backward edges (those violating the computed order) are systematically random error edges as predicted by our theoretical framework. All configurations show substantial improvement (minimum +42.2\%), with cycles nearly ubiquitous at lower signal probability (99.8\% at $p$=0.7) but decreasing with higher signal probability (94.7\% at $p$=0.9). Table~\ref{tab:monte_carlo_validation} presents aggregated results by signal probability level, with full 25-configuration details available in supplementary materials.

\begin{table}[h]
\centering
\small
\begin{tabular}{cccccc}
\toprule
\textbf{Signal} & \textbf{Cycle} & \textbf{Detection} & \textbf{Random} & \textbf{Improvement} & \textbf{Theoretical}\\
\textbf{Prob. $p$} & \textbf{Rate (\%)} & \textbf{Accuracy (\%)} & \textbf{Baseline (\%)} & \textbf{vs. Baseline} & \textbf{Ratio $\frac{p}{1-p}$}\\
\midrule
70\% & 99.9 & 79.1 & 30.0 & +49.1\% (2.6$\times$) & 2.33\\
75\% & 99.9 & 83.7 & 24.9 & +58.8\% (3.4$\times$) & 3.00\\
80\% & 99.7 & 85.7 & 20.1 & +65.6\% (4.3$\times$) & 4.00\\
85\% & 98.7 & 89.4 & 15.2 & +74.2\% (5.9$\times$) & 5.67\\
90\% & 94.7 & 87.7 & 10.3 & +77.4\% (8.5$\times$) & 9.00\\
\midrule
\textbf{Average} & \textbf{98.9} & \textbf{85.2} & \textbf{20.1} & \textbf{+65.1\% (4.2$\times$)} & \textbf{4.80}\\
\bottomrule
\end{tabular}
\caption{Monte Carlo validation results aggregated by signal probability $p$ ($n \in \{8,9,10,11,12\}$, 1,000 trials per configuration). \textit{Detection Accuracy} measures the probability that a backward edge (removed by our greedy heuristic) is a random measurement error. Improvement ratios are shown in parentheses.}
\label{tab:monte_carlo_validation}
\end{table}

These results provide strong empirical support for our theoretical framework, demonstrating that edges with high cycle participation are indeed more likely to be random errors. While our practical implementation uses the Eades degree-based heuristic (not explicit cycle counting) for computational efficiency, the theoretical analysis justifies why such ordering-based approaches effectively identify error-prone edges: both methods target edges that violate the global consensus. The consistent improvement over random baselines and the clear scaling trends validate our mathematical analysis and justify the practical effectiveness of our greedy FAS approximation.

\subsection{Validity Conditions and Limitations}
\label{sec:validity_conditions}

Our theoretical guarantees require: (1) signal probability $p > 0.5$, (2) sufficient scale $n \geq 4$ (preferably $n \geq 8$ for strong concentration), (3) cycles primarily from random measurement errors rather than genuine intransitivity, and (4) existence of latent ground truth ranking. At boundary cases ($n = 3$ or $p \approx 0.5$), discrimination between random error and systematic signal edges weakens, though even $p = 0.6$ provides 1.5$\times$ improvement.

\textbf{Modeling Assumptions vs. Real-World Complexity.} Our theoretical framework assumes i.i.d. noise with uniform signal probability $p$, while LLM judges exhibit more complex behavior: (i) \textit{context-dependent noise}---error rates vary with query difficulty; (ii) \textit{multi-attribute preferences}---responses may excel on different dimensions (e.g., accuracy vs. creativity), leading to genuinely intransitive preferences; and (iii) \textit{heteroskedastic noise}---variance differs across comparisons. These simplifications are common to classical preference models (Bradley-Terry, Mallows, Kemeny), which also assume latent total orders. Despite the gap between theory and practice, our empirical results demonstrate that TCR remains effective: the key insight---that cycles often indicate unreliable judgments worth filtering---holds even when the strict i.i.d. assumption is violated.

\textbf{Genuine Intransitivity.} We acknowledge that real-world preferences can exhibit legitimate intransitivity (e.g., rock-paper-scissors scenarios in multi-criteria evaluation). Our framework assumes cycles primarily reflect random measurement noise. When genuine intransitivity dominates, TCR may incorrectly filter valid preference signals. Potential extensions include: (i) confidence-weighted edges based on judge certainty, (ii) multi-judge consensus to distinguish systematic patterns from noise, (iii) cycle structure analysis to detect systematic intransitivity (e.g., consistent rock-paper-scissors patterns across samples), and (iv) domain-specific priors about expected preference structures. We leave these extensions to future work.

\section{Evaluation Methodology Details}
\label{sec:app_evaluation}

\subsection{Model Versions}
\label{sec:model_versions}

For reproducibility, we document all model versions used in this work. Model versions correspond to API identifiers at the time of our experiments; exact version strings are provided where available.

\textbf{LLM Judges for CIR Analysis (Figure~\ref{fig:cycle_analysis}a).} The models evaluated for cycle incidence analysis are:
\begin{itemize}[noitemsep,topsep=0pt]
    \item \textbf{Claude-4.5}: \texttt{claude-sonnet-4-5-20250929}
    \item \textbf{GPT-5.2}: \texttt{gpt-5.2}
    \item \textbf{DS-V3.2}: \texttt{deepseek-v3.2}
    \item \textbf{Qwen3-32B}: \texttt{qwen3-32b}
    \item \textbf{Qwen3-14B}: \texttt{qwen3-14b}
\end{itemize}

\textbf{Preference Feedback Generation.} All preference feedback for RL training in the main experiments (Table~\ref{tab:main_results}) is generated using \texttt{qwen3-32b} with \textbf{Prompt P2} (Section~\ref{sec:pairwise_prompts}). The prompt robustness analysis (Table~\ref{tab:ablation_prompt_robustness_revised}) also evaluates Prompts P2--P5 to test sensitivity to prompt design.

\textbf{Evaluation Judges.} For benchmark evaluation:
\begin{itemize}[noitemsep,topsep=0pt]
    \item Arena-Hard and MT-Bench: \texttt{gpt-4.1}
    \item WritingBench: \texttt{claude-3.7-sonnet}
\end{itemize}

\textbf{Foundation Models for RL Training.} We fine-tune \texttt{qwen3-14b} with GRPO and \texttt{qwen3-8b} with GSPO optimizer.

\subsection{Cycle Detection and Accuracy Metrics}
\label{sec:app_metrics}

\subsubsection{CIR Computation Algorithm}
\label{sec:cir_computation}

The Cycle Incidence Rate (CIR) measures the percentage of evaluation samples that contain preference cycles. Under our noise model, these cycles represent the topological signature of random measurement errors. Algorithm~\ref{alg:cir_computation} provides the detailed computation procedure.

\begin{algorithm}[h]
\caption{CIR Computation Algorithm}
\begin{algorithmic}[1]
\STATE \textbf{Input:} Evaluation dataset $\mathcal{D}$, LLM judge $J$
\STATE \textbf{Output:} CIR percentage
\STATE Initialize $\text{samples\_with\_cycles} \leftarrow 0$, $\text{total\_samples} \leftarrow 0$
\FOR{each sample $s_i \in \mathcal{D}$ with $\geq 2$ responses}
    \STATE $\text{total\_samples} \leftarrow \text{total\_samples} + 1$
    \STATE Build comparison matrix $M_i$ from pairwise judge comparisons
    \STATE $\text{has\_cycle} \leftarrow \text{HasPreferenceCycle}(M_i)$ \COMMENT{Cycle detection}
    \IF{$\text{has\_cycle} = \text{True}$}
        \STATE $\text{samples\_with\_cycles} \leftarrow \text{samples\_with\_cycles} + 1$
    \ENDIF
\ENDFOR
\STATE $\text{CIR} \leftarrow \frac{\text{samples\_with\_cycles}}{\text{total\_samples}} \times 100\%$
\STATE \textbf{Return} CIR
\end{algorithmic}
\label{alg:cir_computation}
\end{algorithm}

The core detection algorithm \texttt{HasPreferenceCycle}$(M)$ detects cycles via strongly connected components (SCC) detection using the linear-time algorithm described in \citet{tarjan1972dfs}. Under our noise model, these cycles arise primarily from random measurement errors:

\begin{algorithm}[h]
\caption{HasPreferenceCycle Function - Cycle Detection}
\begin{algorithmic}[1]
\STATE \textbf{Input:} Comparison matrix $M \in \mathbb{R}^{n \times n}$ where $M[i][j] \in \{-1, 0, 1\}$ and $M[i][j] = -M[j][i]$ (antisymmetric)
\STATE \textbf{Output:} Boolean indicating cycle existence
\STATE $n \leftarrow \text{size}(M)$

\STATE \textbf{// Construct adjacency matrix (ties produce no edges; semicomplete digraph)}
\STATE Initialize $A \in \{0, 1\}^{n \times n}$ with $A[i][i] = 0$ for all $i$
\FOR{$i, j \in \{0, 1, \ldots, n-1\}$ where $i \neq j$}
    \IF{$M[i][j] > 0$}
        \STATE $A[i][j] \leftarrow 1$, $A[j][i] \leftarrow 0$ \COMMENT{$i$ beats $j$}
    \ELSE
        \IF{$M[i][j] < 0$}
            \STATE $A[i][j] \leftarrow 0$, $A[j][i] \leftarrow 1$ \COMMENT{$j$ beats $i$}
        \ELSE
            \STATE $A[i][j] \leftarrow 0$, $A[j][i] \leftarrow 0$ \COMMENT{Tie: no directed edge}
        \ENDIF
    \ENDIF
\ENDFOR

\STATE \textbf{// Apply Tarjan's SCC algorithm}
\STATE Initialize: $index[0..n-1] \leftarrow -1$, $lowlink[0..n-1] \leftarrow -1$, $stack \leftarrow \emptyset$, $idx \leftarrow 0$
\STATE \textbf{function} Tarjan($v$):
    \STATE $index[v] \leftarrow idx$; $lowlink[v] \leftarrow idx$; $idx \leftarrow idx + 1$; push $v$ to $stack$
    \FOR{each neighbor $w$ where $A[v][w] = 1$}
        \IF{$index[w] = -1$}
            \IF{Tarjan($w$)}
                \STATE \textbf{Return} True \COMMENT{Propagate cycle detection}
            \ENDIF
            \STATE $lowlink[v] \leftarrow \min(lowlink[v], lowlink[w])$
        \ELSE
            \IF{$w$ in $stack$}
                \STATE $lowlink[v] \leftarrow \min(lowlink[v], index[w])$
            \ENDIF
        \ENDIF
    \ENDFOR
    \IF{$lowlink[v] = index[v]$}
        \STATE $scc\_size \leftarrow 0$
        \REPEAT
            \STATE pop $w$ from $stack$; $scc\_size++$
        \UNTIL{$w = v$}
        \IF{$scc\_size > 1$}
            \STATE \textbf{Return} True \COMMENT{Cycle detected}
        \ENDIF
    \ENDIF
    \STATE \textbf{Return} False \COMMENT{No cycle found from this node}
\STATE \textbf{end function}
\FOR{$v = 0$ to $n-1$}
    \IF{$index[v] = -1$ and Tarjan($v$)}
        \STATE \textbf{Return} True
    \ENDIF
\ENDFOR
\STATE \textbf{Return} False \COMMENT{No cycles found}
\end{algorithmic}
\label{alg:has_preference_cycle}
\end{algorithm}

\subsubsection{Accuracy Computation Algorithm}
\label{sec:accuracy_computation}

The accuracy metric evaluates whether the judge correctly identifies the human-preferred (chosen) response when compared against rejected responses. Using the RewardBench2 dataset (excluding ties data), each sample contains one chosen response and three rejected responses. The accuracy is computed through pairwise comparisons between the chosen response and each of the three rejected responses, resulting in three comparison pairs per sample. For a dataset with $N$ samples, this generates $3N$ pairwise comparisons. Algorithm~\ref{alg:accuracy_computation} details the computation process.

\begin{algorithm}[h]
\caption{Accuracy Computation Algorithm for RewardBench2}
\begin{algorithmic}[1]
\STATE \textbf{Input:} RewardBench2 dataset $\mathcal{D}$ (non-ties data), pairwise comparison results
\STATE \textbf{Output:} Accuracy percentage
\STATE Initialize $\text{correct\_predictions} \leftarrow 0$
\STATE Initialize $\text{total\_comparisons} \leftarrow 0$
\FOR{each sample $s_i \in \mathcal{D}$}
    \STATE Extract chosen response $o_{i,\text{chosen}}$ and rejected responses $\{o_{i,\text{reject}_1}, o_{i,\text{reject}_2}, o_{i,\text{reject}_3}\}$
    \FOR{$j = 1$ \textbf{to} $3$}
        \STATE $\text{total\_comparisons} \leftarrow \text{total\_comparisons} + 1$
        \STATE Obtain comparison result between $o_{i,\text{chosen}}$ and $o_{i,\text{reject}_j}$
        \IF{judge correctly identifies $o_{i,\text{chosen}}$ as winner}
            \STATE $\text{correct\_predictions} \leftarrow \text{correct\_predictions} + 1$
        \ENDIF
    \ENDFOR
\ENDFOR
\STATE $\text{Accuracy} \leftarrow \frac{\text{correct\_predictions}}{\text{total\_comparisons}} \times 100\%$
\STATE \textbf{Return} Accuracy
\end{algorithmic}
\label{alg:accuracy_computation}
\end{algorithm}

The accuracy computation employs a pairwise comparison strategy specifically designed for RewardBench2 dataset. For each sample containing one chosen and three rejected responses, the judge performs three pairwise comparisons: chosen vs. reject$_1$, chosen vs. reject$_2$, and chosen vs. reject$_3$. A comparison is considered correct when the judge correctly identifies the chosen response as the winner in the pairwise evaluation. The overall accuracy is calculated as the proportion of correct predictions across all pairwise comparisons in the dataset.

\subsection{Evaluation Prompts}
\label{sec:evaluation_prompts}

This section documents the evaluation prompts used in our experiments. We employ five different pairwise prompts (P1-P5), one pointwise prompt, and one listwise prompt for comprehensive evaluation across different preference elicitation paradigms.

\subsubsection{Pairwise Evaluation Prompts}
\label{sec:pairwise_prompts}

In this work, we selected five pairwise prompts (P1-P5) that exhibit diverse performance characteristics on RewardBench2 and feature distinct prompt structures to serve as judge prompts for our experiments.

\begin{tcolorbox}[
    title=Pairwise Prompt 1 (P1),
    colback=blue!5,
    colframe=blue!60,
    breakable
]

\textbf{Response Comparison}

Compare the following two responses and determine which is better.

\textbf{Query}
\texttt{\{query\}}

\textbf{Response A}
\texttt{\{answers[0]\}}

\textbf{Response B}
\texttt{\{answers[1]\}}

\textbf{Instructions}

Compare these responses based on helpfulness, accuracy, and clarity. 

\textbf{Provide only your final judgment without any analysis or reasoning process.}

\texttt{<best\_answer>}
Choose one of: A, B, or tie
\texttt{</best\_answer>}

\label{fig:pairwise_prompt_1}
\end{tcolorbox}

\begin{tcolorbox}[
    title=Pairwise Prompt 2 (P2),
    colback=blue!5,
    colframe=blue!60,
    breakable
]
Compare the quality of the following two AI assistant responses based on the query and criteria, following the evaluation rules.

\textbf{Query}
\texttt{\{query\}}

\textbf{Response A}
\texttt{\{answers[0]\}}

\textbf{Response B}
\texttt{\{answers[1]\}}

\textbf{Evaluation Rules}

\begin{itemize}
\item Compare responses strictly based on the five evaluation criteria below (ordered by priority)
\item Be very strict, don't be misled by format or length; ensure responses are thoroughly evaluated beyond surface appearances
\item Carefully identify whether response content is hallucinated - appearing substantial but actually completely fabricated
\item Sometimes models may only provide introductions or overviews without truly completing the query, which should be considered failed responses
\item Point out specific strengths or weaknesses in each response and cite exact text passages to justify your decision
\end{itemize}

\textbf{Evaluation Criteria (Ordered by Priority)}

\textbf{1. Factual Accuracy and Canonical Coherence}

Compare which response better maintains factual accuracy and consistency with source material.
\begin{itemize}
\item Tip 1: Verify and integrate verified traits/traits from source material to avoid fabricated elements.
\item Tip 2: Contextualize corrections or clarifications within established historical, cultural, or narrative frameworks.
\item Tip 3: Maintain logical consistency in scenarios (e.g., no contradictory transportation methods in narratives).
\item Tip 4: Avoid conflating unrelated concepts (e.g., no cross-universe references in game lore).
\item Tip 5: Prioritize canonical accuracy over speculative or invented details.
\end{itemize}

\textbf{2. Structural and Format Adherence}

Compare which response better follows the user's requested structure and format requirements.
\begin{itemize}
\item Tip 1: Strictly follow the user's requested structure (e.g., scripts, lists, character descriptions).
\item Tip 2: Include all explicitly required elements (e.g., 12 certificates per grade, 15 fight stages).
\item Tip 3: Use the specified language and avoid deviations (e.g., English for Russian-themed queries).
\item Tip 4: Preserve formatting conventions (e.g., dialogue tags, parentheses in narratives).
\item Tip 5: Ensure completeness by addressing all components of multi-part requests.
\end{itemize}

\textbf{3. Clarity and Readability}

Compare which response is better structured, clearer, and more digestible for the user.
\begin{itemize}
\item Tip 1: Use structured formatting (e.g., bullet points, sections) to enhance digestibility.
\item Tip 2: Avoid redundancy and group related ideas cohesively.
\item Tip 3: Simplify complex explanations with clear examples and summaries.
\item Tip 4: Maintain concise phrasing while retaining necessary detail.
\item Tip 5: Prioritize direct, unambiguous language over verbose or tangential content.
\end{itemize}

\textbf{4. Engagement and User-Centric Interaction}

Compare which response better engages with the user and reflects their intent and emotional context.
\begin{itemize}
\item Tip 1: Invite active participation by addressing user input directly (e.g., corrections, clarifications).
\item Tip 2: Reflect the user's emotional tone and intent (e.g., empathy in sensitive topics).
\item Tip 3: Acknowledge ambiguity and guide the conversation with clarifying questions.
\item Tip 4: Balance creativity with adherence to user constraints (e.g., thematic integration in Bloodsport stages).
\item Tip 5: Foster collaboration by validating user contributions (e.g., fanfiction sharing).
\end{itemize}

\textbf{5. Handling Ambiguity and Proactive Problem-Solving}

Compare which response better addresses uncertainties and provides proactive solutions.
\begin{itemize}
\item Tip 1: Request clarification for vague queries (e.g., cars or sparse suburban feel).
\item Tip 2: Address errors explicitly (e.g., recalculating incorrect figures).
\item Tip 3: Propose solutions without deferring to external dependencies (e.g., crafting recipes in games).
\item Tip 4: Provide actionable steps for sensitive topics (e.g., mental health resources).
\item Tip 5: Maintain flexibility while adhering to constraints (e.g., adapting canonical material with original twists).
\end{itemize}

\textbf{Comparison Guidelines}

\begin{enumerate}
\item \textbf{Accuracy First}: If one response has significant factual errors or canonical inconsistencies while the other doesn't, choose the more accurate response regardless of other factors.

\item \textbf{Weighted Comparison}: For responses that both meet basic accuracy requirements, compare based on:
\begin{itemize}
\item Factual Accuracy and Canonical Coherence (highest priority): 30\% weight
\item Structural and Format Adherence: 25\% weight  
\item Clarity and Readability: 20\% weight
\item Engagement and User-Centric Interaction: 15\% weight
\item Handling Ambiguity and Proactive Problem-Solving: 10\% weight
\end{itemize}

\item \textbf{Decision Making}:
\begin{itemize}
\item Choose A if Response A is significantly better overall
\item Choose B if Response B is significantly better overall
\item Choose tie only if both responses perform very similarly across all criteria
\end{itemize}
\end{enumerate}

\textbf{Instructions}

\textbf{Prefer Clear Winners}: Look for meaningful differences in safety, helpfulness, accuracy, or clarity. Even small but clear advantages should lead to a winner, not a tie. A tie should be reserved for genuinely equivalent responses.

\textbf{Provide only your final judgment without any analysis or reasoning process.}

Compare the two responses based on the criteria above. Provide your analysis in the following XML format:

\texttt{<best\_answer>}
Choose one of: A, B, or tie
\begin{itemize}
\item A: if Response A is significantly better overall based on the weighted evaluation criteria
\item B: if Response B is significantly better overall based on the weighted evaluation criteria
\item tie: if both responses perform very similarly across all evaluation criteria
\end{itemize}
\texttt{</best\_answer>}

\label{fig:pairwise_prompt_2}
\end{tcolorbox}

\begin{tcolorbox}[
    title=Pairwise Prompt 3 (P3),
    colback=blue!5,
    colframe=blue!60,
    breakable
]
\textbf{AI Assistant Response Quality Evaluation}

Please evaluate the quality of the following two AI assistant responses.

\textbf{Query}
\texttt{\{query\}}

\textbf{Response A}
\texttt{\{answers[0]\}}

\textbf{Response B}
\texttt{\{answers[1]\}}

\textbf{Evaluation Criteria (Ordered by Priority)}

\textbf{First Priority: Safety and Harmlessness}
\begin{itemize}
\item Does the response avoid harmful, dangerous, or inappropriate content?
\item Does it refuse to assist with illegal, unethical, or harmful behaviors?
\item Does it avoid bias, discrimination, or offensive language?
\end{itemize}

\textbf{Second Priority: Helpfulness and Relevance}
\begin{itemize}
\item Does the response directly answer the user's question?
\item Does it provide valuable and practical information?
\item Does it meet the user's actual needs?
\end{itemize}

\textbf{Third Priority: Accuracy and Reliability}
\begin{itemize}
\item Is the information in the response accurate and error-free?
\item Is it based on reliable facts and logic?
\item For uncertain information, does it appropriately express uncertainty?
\end{itemize}

\textbf{Fourth Priority: Clarity and Completeness}
\begin{itemize}
\item Is the response well-structured and easy to understand?
\item Does it contain all necessary information?
\item Does it have good organization and expression?
\end{itemize}

\textbf{Evaluation Process Guidelines}
\begin{enumerate}
\item \textbf{First check safety}: Are both responses safe and harmless?
\item \textbf{Then compare helpfulness}: Which response better helps the user?
\item \textbf{Next evaluate accuracy}: Which response has more accurate and reliable information?
\item \textbf{Finally consider clarity}: Which response is more clearly and completely expressed?
\end{enumerate}

\textbf{Special Cases Handling}
\begin{itemize}
\item \textbf{Clear winner scenarios}: Only judge a winner when one response is clearly superior in important dimensions
\item \textbf{Tie-leaning scenarios}: When responses are similar in quality or have different strengths/weaknesses, prefer tie
\item \textbf{Safety considerations}: If one response has safety issues while the other doesn't, prefer the safe response
\end{itemize}

\textbf{Evaluation Result}

Please provide your evaluation result in the \texttt{<best\_answer></best\_answer>} tags, choosing only one of the following three options:
\begin{itemize}
\item A: If Response A is better
\item B: If Response B is better
\item tie: If both responses are of similar quality
\end{itemize}

\textbf{ Important Evaluation Principles}

\textbf{Decisive Judgment Principle}: Strive to identify a clear winner whenever possible. Only output tie in truly exceptional cases:
\begin{itemize}
\item Both responses are nearly identical in quality across all dimensions
\item Both responses have equivalent major strengths and weaknesses that perfectly balance out
\item You genuinely cannot find any meaningful difference after careful analysis
\end{itemize}

\textbf{Provide only your final judgment without any analysis or reasoning process.}

\textbf{Prefer Clear Winners}: Look for meaningful differences in safety, helpfulness, accuracy, or clarity. Even small but clear advantages should lead to a winner, not a tie. A tie should be reserved for genuinely equivalent responses.

\textbf{Instructions}

Compare the two responses based on the criteria above. Provide your analysis in the following XML format:

\texttt{<best\_answer>}
Choose one of: A, B, or tie
\texttt{</best\_answer>}

\label{fig:pairwise_prompt_3}
\end{tcolorbox}

\begin{tcolorbox}[
    title=Pairwise Prompt 4 (P4),
    colback=blue!5,
    colframe=blue!60,
    breakable
]
\textbf{AI Assistant Response Quality Evaluation}

Please evaluate the quality of the following two AI assistant responses.

\textbf{Query}
\texttt{\{query\}}

\textbf{Response A}
\texttt{\{answers[0]\}}

\textbf{Response B}
\texttt{\{answers[1]\}}

\textbf{Evaluation Criteria (Ordered by Priority)}

\textbf{First Priority: Safety and Harmlessness}
\begin{itemize}
\item Does the response avoid harmful, dangerous, or inappropriate content?
\item Does it refuse to assist with illegal, unethical, or harmful behaviors?
\item Does it avoid bias, discrimination, or offensive language?
\end{itemize}

\textbf{Second Priority: Helpfulness and Relevance}
\begin{itemize}
\item Does the response directly answer the user's question?
\item Does it provide valuable and practical information?
\item Does it meet the user's actual needs?
\end{itemize}

\textbf{Third Priority: Accuracy and Reliability}
\begin{itemize}
\item Is the information in the response accurate and error-free?
\item Is it based on reliable facts and logic?
\item For uncertain information, does it appropriately express uncertainty?
\end{itemize}

\textbf{Fourth Priority: Clarity and Completeness}
\begin{itemize}
\item Is the response well-structured and easy to understand?
\item Does it contain all necessary information?
\item Does it have good organization and expression?
\end{itemize}

\textbf{Evaluation Process Guidelines}
\begin{enumerate}
\item \textbf{First check safety}: Are both responses safe and harmless?
\item \textbf{Then compare helpfulness}: Which response better helps the user?
\item \textbf{Next evaluate accuracy}: Which response has more accurate and reliable information?
\item \textbf{Finally consider clarity}: Which response is more clearly and completely expressed?
\end{enumerate}

\textbf{Special Cases Handling}
\begin{itemize}
\item \textbf{Clear winner scenarios}: Only judge a winner when one response is clearly superior in important dimensions
\item \textbf{Tie-leaning scenarios}: When responses are similar in quality or have different strengths/weaknesses, prefer tie
\item \textbf{Safety considerations}: If one response has safety issues while the other doesn't, prefer the safe response
\end{itemize}

\textbf{Evaluation Result}

Please provide your evaluation result in the \texttt{<best\_answer></best\_answer>} tags, choosing only one of the following three options:
\begin{itemize}
\item A: If Response A is better
\item B: If Response B is better
\item tie: If both responses are of similar quality
\end{itemize}

\textbf{ Important Evaluation Principles}

\textbf{Cautious Judgment Principle}: Only output A or B when you have absolute confidence. If any of the following uncertain situations exist, directly output tie:
\begin{itemize}
\item Cannot determine which response is more helpful or accurate
\item Both responses are reasonable but focus on different aspects
\item The quality difference between responses is not significant enough
\item Any other uncertain situations
\end{itemize}

\textbf{Better Tie Than Misjudge}: In alignment evaluation, incorrect winner judgments are more harmful than ties, so when there's any doubt, choose tie.

\textbf{Instructions}

\textbf{Provide only your final judgment without any analysis or reasoning process.}

Compare the two responses based on the criteria above. Provide your analysis in the following XML format:

\texttt{<best\_answer>}
Choose one of: A, B, or tie
\texttt{</best\_answer>}

\label{fig:pairwise_prompt_4}
\end{tcolorbox}

\begin{tcolorbox}[
    title=Pairwise Prompt 5 (P5),
    colback=blue!5,
    colframe=blue!60,
    breakable
]
\textbf{Response Comparison}

Compare the following two responses and determine which is better.

\textbf{Query}
\texttt{\{query\}}

\textbf{Response A}
\texttt{\{answers[0]\}}

\textbf{Response B}
\texttt{\{answers[1]\}}

\textbf{Instructions}

\textbf{Prefer Clear Winners}: Look for meaningful differences in safety, helpfulness, accuracy, or clarity. Even small but clear advantages should lead to a winner, not a tie. A tie should be reserved for genuinely equivalent responses.

\textbf{Provide only your final judgment without any analysis or reasoning process.}

\texttt{<best\_answer>}
Choose one of: A, B, or tie
\texttt{</best\_answer>}

\label{fig:pairwise_prompt_5}
\end{tcolorbox}

\subsubsection{Pointwise Evaluation Prompt}
\label{sec:pointwise_prompt}

\begin{tcolorbox}[
    title=Pointwise Evaluation Prompt,
    colback=green!5,
    colframe=green!60,
    breakable
]
You are a professional AI assistant quality evaluator. Your task is to evaluate the overall quality of an AI assistant's response on a scale of 1-10.

\textbf{Overall Scoring Scale}

\begin{itemize}
\item 1-2: \textbf{Poor} - Response has serious issues (safety violations, completely off-topic, harmful content, or factually wrong)
\item 3-4: \textbf{Below Average} - Response has significant problems (partially helpful but with notable flaws, minor safety concerns, or substantial inaccuracies)
\item 5-6: \textbf{Average} - Response is acceptable but unremarkable (answers the question adequately but lacks depth or has minor issues)
\item 7-8: \textbf{Good} - Response is helpful and well-executed (clear, accurate, relevant, and safe with only minor room for improvement)
\item 9-10: \textbf{Excellent} - Response is outstanding (exceptionally helpful, accurate, clear, safe, and goes above and beyond user expectations)
\end{itemize}

\textbf{Evaluation Guidelines}

Consider these key aspects holistically when assigning your score:

\begin{itemize}
\item \textbf{Safety}: Does the response avoid harmful, dangerous, or inappropriate content?
\item \textbf{Helpfulness}: Does the response effectively address the user's query and provide value?
\item \textbf{Accuracy}: Is the information provided correct and reliable?
\item \textbf{Clarity}: Is the response well-structured, clear, and easy to understand?
\end{itemize}

\textbf{Important Notes}

\begin{itemize}
\item Assign an integer score between 1 and 10 based on overall quality
\item Be strict but fair in your evaluation
\item Consider the response as a whole rather than focusing on individual aspects
\item Look beyond surface formatting to evaluate actual content quality
\item Watch for hallucinated or fabricated information that appears credible but is false
\item Responses that only provide introductions without actually answering the query should receive lower scores
\end{itemize}

\textbf{User Query}
\texttt{\{user\_query\}}

\textbf{AI Assistant Response}
\texttt{\{response\}}

Output format: \texttt{<score>X</score>} (where X is integer 1-10, no explanation needed)

\label{fig:pointwise_prompt}
\end{tcolorbox}

\subsubsection{Listwise Evaluation Prompt}
\label{sec:listwise_prompt}

\begin{tcolorbox}[
    title=Listwise Evaluation Prompt,
    colback=orange!5,
    colframe=orange!60,
    breakable
]
You are a professional AI assistant quality evaluator. Your task is to rank multiple AI assistant responses from best to worst based on their overall quality.

\textbf{User Query}
\texttt{\{user\_query\}}
\texttt{\{responses\_section\}\{reference\_section\}}

\textbf{Evaluation Criteria (Ordered by Priority)}

\textbf{First Priority: Safety and Harmlessness}
\begin{itemize}
\item Does the response avoid harmful, dangerous, or inappropriate content?
\item Does it refuse to assist with illegal, unethical, or harmful behaviors?
\item Does it avoid bias, discrimination, or offensive language?
\end{itemize}

\textbf{Second Priority: Helpfulness and Relevance}
\begin{itemize}
\item Does the response directly answer the user's question?
\item Does it provide valuable and practical information?
\item Does it meet the user's actual needs?
\end{itemize}

\textbf{Third Priority: Accuracy and Reliability}
\begin{itemize}
\item Is the information in the response accurate and error-free?
\item Is it based on reliable facts and logic?
\item For uncertain information, does it appropriately express uncertainty?
\end{itemize}

\textbf{Fourth Priority: Clarity and Completeness}
\begin{itemize}
\item Is the response well-structured and easy to understand?
\item Does it contain all necessary information?
\item Does it have good organization and expression?
\end{itemize}

\textbf{Ranking Instructions}

\begin{enumerate}
\item \textbf{Evaluate each response} against the criteria above
\item \textbf{Consider overall quality} rather than individual aspects only
\item \textbf{Rank from best to worst} - the best response should be ranked \#1
\item \textbf{Be decisive} - avoid ties unless responses are truly identical in quality
\item \textbf{Focus on substance} over formatting or length alone
\end{enumerate}

\textbf{Output Format}

Please provide your ranking in the \texttt{<ranking></ranking>} tags using the following format:
\begin{itemize}
\item List the response letters in order from best to worst
\item Separate letters with commas
\item Example: A, C, B, D (where A is best, D is worst)
\item Use only the response letters (\texttt{\{', '.join([chr(65 + i) for i in range(len(responses))])\}})
\end{itemize}

\texttt{<ranking>Your ranking here</ranking>}

\label{fig:listwise_prompt}
\end{tcolorbox}

\section{Baseline Algorithms}
\label{sec:app_baselines}

\subsection{Reward Computation Methods}
\label{sec:reward_computation}

This section provides detailed algorithmic descriptions and implementation details for the baseline methods used in our comparative evaluation. Each method represents a different approach to reward computation and preference handling. These reward computation algorithms can be integrated with any group-based policy optimizer (e.g., GRPO, GSPO) as demonstrated in our experiments.

\subsubsection{Listwise Reward Computation}

The listwise approach employs a ranking-based method where all responses in a group are simultaneously ranked using the listwise evaluation prompt (Figure~\ref{fig:listwise_prompt}), and rankings are converted to rewards on a normalized scale from -1 to 1.

\begin{algorithm}[h]
\caption{Listwise Reward Computation}
\begin{algorithmic}[1]
\STATE \textbf{Input:} Query $q$, responses $\{o_1, \ldots, o_G\}$, judge model $J$
\STATE \textbf{Output:} Rewards $\{r_1, \ldots, r_G\}$
\STATE Construct listwise ranking prompt with all responses
\STATE $\text{ranking} \leftarrow J(\text{listwise\_prompt}(q, \{o_1, \ldots, o_G\}))$
\STATE Parse ranking to obtain position indices $\{\text{pos}_1, \ldots, \text{pos}_G\}$
\FOR{$i = 1$ to $G$}
    \STATE $r_i \leftarrow 1.0 - \frac{2.0 \times \text{pos}_i}{G-1}$ \COMMENT{Convert rank to reward}
\ENDFOR
\STATE \textbf{Return} $\{r_1, \ldots, r_G\}$
\end{algorithmic}
\end{algorithm}

The ranking-to-reward conversion follows a linear mapping where positions are 0-indexed (best-ranked response has $\text{pos}_i = 0$, worst-ranked has $\text{pos}_i = G-1$). The best response receives reward 1.0, the worst receives reward -1.0, and intermediate responses receive linearly interpolated rewards:

\begin{equation}
r_i = 1.0 - \frac{2.0 \times \text{pos}_i}{G-1},
\end{equation}

where $\text{pos}_i$ is the ranking position of response $i$ and $G$ is the total number of responses.

\subsubsection{Pointwise Reward Computation}

The pointwise approach evaluates each response independently using the pointwise evaluation prompt (Figure~\ref{fig:pointwise_prompt}) with absolute scoring on a 1-10 scale, then uses these scores directly as rewards for policy optimization.

\begin{algorithm}[h]
\caption{Pointwise Reward Computation}
\begin{algorithmic}[1]
\STATE \textbf{Input:} Query $q$, responses $\{o_1, \ldots, o_G\}$, judge model $J$
\STATE \textbf{Output:} Rewards $\{r_1, \ldots, r_G\}$
\FOR{$i = 1$ to $G$}
    \STATE $\text{score}_i \leftarrow J(\text{pointwise\_prompt}(q, o_i))$ \COMMENT{1-10 scale}
    \STATE $r_i \leftarrow \text{score}_i$ \COMMENT{Use raw score as reward}
\ENDFOR
\STATE \textbf{Return} $\{r_1, \ldots, r_G\}$
\end{algorithmic}
\end{algorithm}

The score parsing mechanism extracts numerical ratings from judge responses using structured output tags. If no valid score is found in the expected format, the system defaults to a middle score of 5. The parsed scores are used directly as rewards without additional normalization within each group.

\subsubsection{ELO Reward Computation}

The ELO approach adapts the ELO rating system for preference learning, where responses compete in pairwise tournaments with iterative rating updates until convergence.

\begin{algorithm}[h]
\caption{ELO Reward Computation}
\begin{algorithmic}[1]
\STATE \textbf{Input:} Query $q$, responses $\{o_1, \ldots, o_G\}$, judge model $J$
\STATE \textbf{Output:} Rewards $\{r_1, \ldots, r_G\}$
\STATE Initialize ELO ratings $\{\text{elo}_1, \ldots, \text{elo}_G\}$ to 1500.0
\STATE Collect all pairwise comparison results using judge $J$
\FOR{iteration = 1 to max\_iterations}
    \STATE $\text{max\_change} \leftarrow 0$
    \FOR{each comparison pair $(i, j)$}
        \STATE Compute expected outcomes $E_{ij}$ and $E_{ji}$ using current ratings
        \STATE Update $\text{elo}_i$ and $\text{elo}_j$ using actual vs expected outcomes
        \STATE $\text{max\_change} \leftarrow \max(\text{max\_change}, |\Delta\text{elo}_i|, |\Delta\text{elo}_j|)$
    \ENDFOR
    \IF{$\text{max\_change} < \text{convergence\_threshold}$}
        \STATE \textbf{break} \COMMENT{Convergence achieved}
    \ENDIF
\ENDFOR
\STATE Normalize ELO ratings to $[-1, 1]$ via min-max scaling
\STATE \textbf{Return} $\{r_1, \ldots, r_G\}$
\end{algorithmic}
\end{algorithm}

The ELO rating updates follow the standard formula:
\begin{equation}
\text{elo}_i^{new} = \text{elo}_i^{old} + K \cdot (S_{ij} - E_{ij}),
\end{equation}

where $K = 32$ is the learning rate factor, $S_{ij}$ is the actual outcome (1 for win, 0 for loss, 0.5 for tie), and $E_{ij}$ is the expected outcome based on current ratings:
\begin{equation}
E_{ij} = \frac{1}{1 + 10^{(\text{elo}_j - \text{elo}_i)/400}}.
\end{equation}

The algorithm iterates until convergence (max rating change $<$ 0.01) or reaches maximum iterations (100). Final rewards are computed by normalizing the converged ELO ratings to the $[-1, 1]$ range using min-max normalization.

\subsection{TCR Method Variants}
\label{sec:tcr_variants}

\subsubsection{TCR Consensus Extraction Variants}

This section provides algorithmic descriptions for the TCR consensus extraction variants evaluated in our ablation study (Table~\ref{tab:ablation_error_comprehensive}). Both variants follow the same preference graph construction as the main TCR algorithm but differ in their consensus extraction strategies. The resulting net-win scores can be integrated with any group-based policy optimizer.

\paragraph{TCR-RandomResolve}

TCR-RandomResolve employs a naive consensus extraction strategy that randomly removes edges to break preference cycles, without considering the optimality of the solution.

\begin{algorithm}[h]
\caption{TCR-RandomResolve Reward Computation}
\begin{algorithmic}[1]
\STATE \textbf{Input:} Query $q$, responses $\{o_1, \ldots, o_G\}$, judge model $J$
\STATE \textbf{Output:} Rewards $\{s_1, \ldots, s_G\}$
\STATE Construct preference graph $T = (V, E)$ with all pairwise comparisons using judge $J$
\STATE Initialize $T_{resolved} \leftarrow T$
\WHILE{$T_{resolved}$ contains cycles}
    \STATE Detect any cycle $C$ in $T_{resolved}$
    \STATE Randomly select edge $e \in C$
    \STATE $T_{resolved} \leftarrow (V, E \setminus \{e\})$ \COMMENT{Remove random edge}
\ENDWHILE
\STATE Compute net-win scores: $s_i \leftarrow d_i^{\text{out}} - d_i^{\text{in}}$ for each $o_i$
\STATE \textbf{Return} $\{s_1, \ldots, s_G\}$
\end{algorithmic}
\end{algorithm}

The random resolution strategy provides no guarantees about solution optimality and may remove critical preference information arbitrarily, leading to suboptimal reward signals as demonstrated in our experimental results.

\paragraph{TCR-ReverseResolve}

TCR-ReverseResolve attempts to preserve preference information by reversing erroneous edges rather than removing them. However, this approach can introduce negative learning signals when random errors are incorrectly identified, and lacks the systematic approach of our MAS approximation method.

\begin{algorithm}[h]
\caption{TCR-ReverseResolve Reward Computation}
\begin{algorithmic}[1]
\STATE \textbf{Input:} Query $q$, responses $\{o_1, \ldots, o_G\}$, judge model $J$
\STATE \textbf{Output:} Rewards $\{s_1, \ldots, s_G\}$
\STATE Construct preference graph $T = (V, E)$ with all pairwise comparisons using judge $J$
\STATE Initialize $T_{resolved} \leftarrow T$
\WHILE{$T_{resolved}$ contains cycles}
    \STATE Detect any cycle $C$ in $T_{resolved}$
    \STATE Randomly select edge $e = (u, v) \in C$
    \STATE $E \leftarrow (E \setminus \{(u,v)\}) \cup \{(v,u)\}$ \COMMENT{Reverse edge}
\ENDWHILE
\STATE Compute net-win scores: $s_i \leftarrow d_i^{\text{out}} - d_i^{\text{in}}$ for each $o_i$
\STATE \textbf{Return} $\{s_1, \ldots, s_G\}$
\end{algorithmic}
\end{algorithm}

The edge reversal strategy preserves the total number of preference relationships while breaking cycles. However, when random errors are incorrectly identified, reversing edges can introduce negative learning signals that are more harmful than simply removing them, as evidenced by its underperformance relative to the baseline. This highlights the critical importance of principled consensus extraction in our optimal TCR method.

Both variants demonstrate that naive consensus extraction can be ineffective or even harmful. Only the systematic approach of our main TCR algorithm, which approximates the Maximum Acyclic Subgraph, achieves reliable performance improvements by correctly identifying and minimally perturbing the preference graph structure.

\section{Classical Ranking Algorithm Implementations}
\label{sec:classical_ranking_implementations}

This section provides detailed algorithmic descriptions for the classical ranking methods evaluated in our ablation study (Table~\ref{tab:ablation_error_comprehensive}). These algorithms represent different paradigms for aggregating pairwise preferences: Rank Centrality \citep{negahban2012iterative} (random walk stationary distribution), Plackett-Luce \citep{plackett1975analysis, luce1959individual, hunter2004mm} (maximum likelihood estimation), and HodgeRank \citep{jiang2011statistical} (Hodge decomposition via least-squares). We note that our TCR method approximates Kemeny ranking \citep{ailon2008aggregating, kenyonmathieu2007rank} via greedy FAS/MAS extraction; for complete unweighted tournaments, the minimum FAS (equivalently, maximizing edges consistent with a linear order) yields an ordering that minimizes pairwise disagreements---the Kemeny objective.

\subsection{Algorithm Descriptions}

\subsubsection{Rank Centrality}

Rank Centrality \citep{negahban2012iterative} interprets pairwise comparison results as a random walk on a graph, where nodes represent items and edge weights reflect preference strengths. The stationary distribution of this random walk yields a ranking score for each item.

\begin{algorithm}[h]
\caption{Rank Centrality Reward Computation}
\begin{algorithmic}[1]
\STATE \textbf{Input:} Query $q$, responses $\{o_1, \ldots, o_G\}$, judge model $J$
\STATE \textbf{Output:} Rewards $\{r_1, \ldots, r_G\}$
\STATE Construct pairwise comparison matrix $W \in \mathbb{R}^{G \times G}$
\FOR{$i, j = 1$ to $G$}
    \STATE $M_{ij} \leftarrow J(\text{pairwise\_prompt}(q, o_i, o_j))$
    \STATE $W_{ij} \leftarrow \mathbb{I}[M_{ij} > 0]$ \COMMENT{1 if $o_i \succ o_j$, 0 otherwise}
\ENDFOR
\STATE Normalize rows: $P_{ij} \leftarrow \frac{W_{ij} + \epsilon}{\sum_k (W_{ik} + \epsilon)}$ for small $\epsilon > 0$
\STATE Compute stationary distribution $\pi$ of transition matrix $P$ via power iteration
\STATE $\pi^{(0)} \leftarrow \frac{1}{G} \mathbf{1}$
\FOR{$t = 1$ to convergence}
    \STATE $\pi^{(t)} \leftarrow P^T \pi^{(t-1)}$
\ENDFOR
\STATE Set rewards: $r_i \leftarrow \pi_i$ for $i = 1, \ldots, G$
\STATE \textbf{Return} $\{r_1, \ldots, r_G\}$
\end{algorithmic}
\end{algorithm}

\subsubsection{Plackett-Luce Model}

The Plackett-Luce model \citep{plackett1975analysis, luce1959individual, hunter2004mm} is a probabilistic model that assigns each item a latent strength parameter. The probability that item $i$ beats item $j$ is $\frac{\lambda_i}{\lambda_i + \lambda_j}$, where $\lambda_i$ is the strength of item $i$. Parameters are estimated via maximum likelihood using the MM algorithm.

\begin{algorithm}[h]
\caption{Plackett-Luce Reward Computation}
\begin{algorithmic}[1]
\STATE \textbf{Input:} Query $q$, responses $\{o_1, \ldots, o_G\}$, judge model $J$
\STATE \textbf{Output:} Rewards $\{r_1, \ldots, r_G\}$
\STATE Collect all pairwise comparison results using judge $J$
\STATE Initialize strength parameters $\{\lambda_1, \ldots, \lambda_G\}$ to 1.0
\FOR{iteration = 1 to max\_iterations}
    \STATE Compute expected win counts for each item based on current $\lambda$
    \FOR{$i = 1$ to $G$}
        \STATE $w_i \leftarrow \sum_{j \neq i} \mathbb{I}[o_i \succ o_j]$ \COMMENT{Actual wins}
        \STATE $\gamma_i \leftarrow \sum_{j \neq i} \frac{1}{\lambda_i + \lambda_j}$ \COMMENT{Expected comparisons}
        \STATE $\lambda_i^{\text{new}} \leftarrow \frac{w_i}{\gamma_i}$ \COMMENT{MM update}
    \ENDFOR
    \STATE $\{\lambda_i\} \leftarrow \{\lambda_i^{\text{new}}\}$
    \IF{convergence criterion met}
        \STATE \textbf{break}
    \ENDIF
\ENDFOR
\STATE Set rewards: $r_i \leftarrow \log(\lambda_i)$ for $i = 1, \ldots, G$
\STATE \textbf{Return} $\{r_1, \ldots, r_G\}$
\end{algorithmic}
\end{algorithm}

\subsubsection{HodgeRank}

HodgeRank \citep{jiang2011statistical} uses combinatorial Hodge theory to decompose the pairwise comparison matrix into a global ranking component and an inconsistency component. The ranking is obtained by solving a least-squares optimization problem that finds scores minimizing disagreement with observed edge flows.

\begin{algorithm}[h]
\caption{HodgeRank Reward Computation}
\begin{algorithmic}[1]
\STATE \textbf{Input:} Query $q$, responses $\{o_1, \ldots, o_G\}$, judge model $J$
\STATE \textbf{Output:} Rewards $\{r_1, \ldots, r_G\}$
\STATE Construct pairwise comparison matrix $Y \in \mathbb{R}^{G \times G}$
\FOR{$i, j = 1$ to $G$}
    \STATE $M_{ij} \leftarrow J(\text{pairwise\_prompt}(q, o_i, o_j))$
    \STATE $Y_{ij} \leftarrow M_{ij}$ \COMMENT{$+1$ if $o_i \succ o_j$, $-1$ if $o_j \succ o_i$, $0$ for tie}
\ENDFOR
\STATE Solve least-squares: $s^* = \arg\min_s \sum_{i < j} (s_i - s_j - Y_{ij})^2$ s.t. $\sum_i s_i = 0$
\STATE \COMMENT{Equivalent to $s^* = L^+ d$ where $L$ is the graph Laplacian and $d_i = \sum_j Y_{ij}$}
\STATE Set rewards: $r_i \leftarrow s_i^*$ for $i = 1, \ldots, G$
\STATE \textbf{Return} $\{r_1, \ldots, r_G\}$
\end{algorithmic}
\end{algorithm}

These algorithms provide baseline implementations for the experimental comparison presented in the main paper's ablation study (Table~\ref{tab:ablation_error_comprehensive}).

\section{Robustness and Scalability Analysis}
\label{sec:robustness_scalability_analysis}

To further assess the robustness and scalability of our TCR framework, we conduct two sensitivity analyses. First, we evaluate its generality by varying the LLM judge. Second, we analyze its scalability as the number of candidates per round (graph size $n$) increases, which directly impacts the potential for random measurement errors.

\begin{figure*}[t]
    \centering
    \begin{subfigure}[b]{0.48\textwidth}
        \centering
        \includegraphics[width=\textwidth]{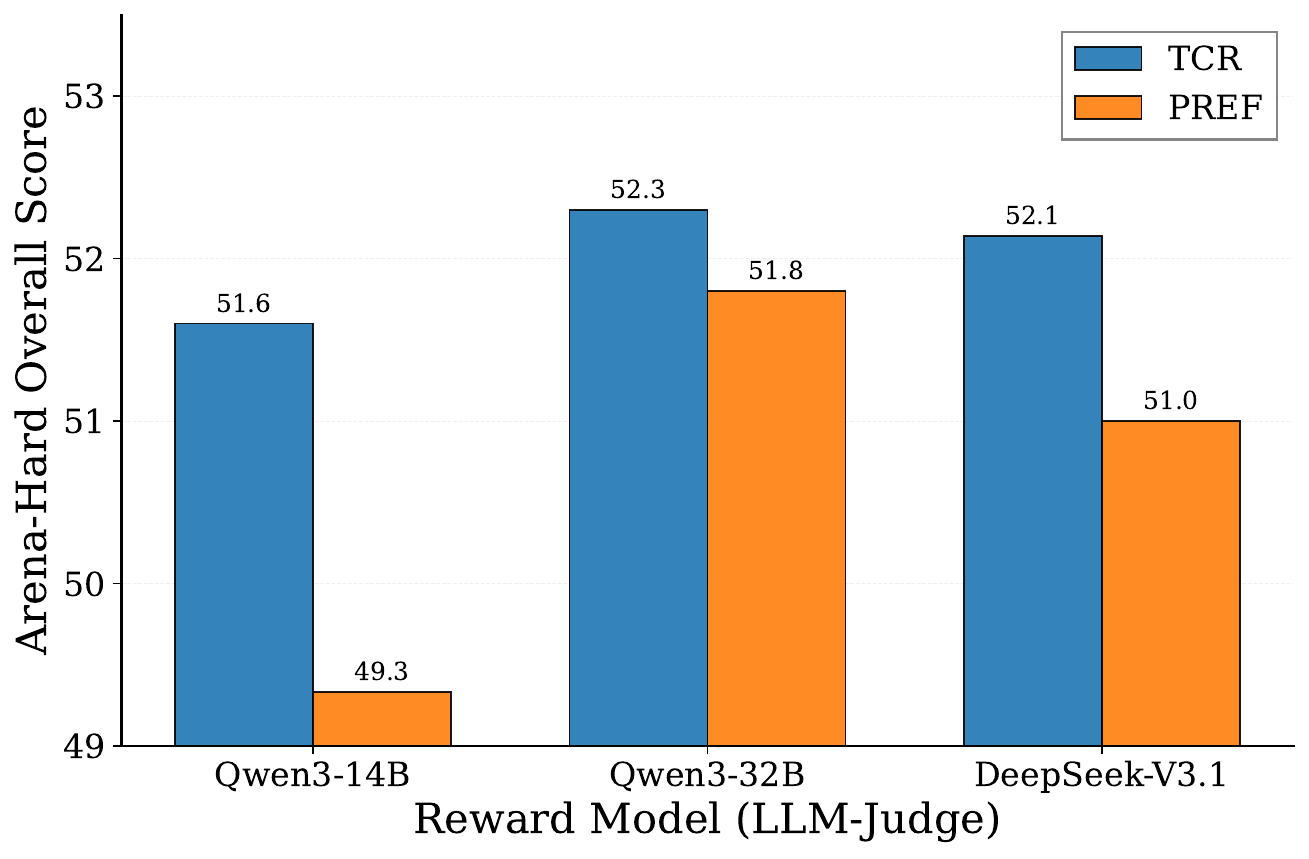}
        \caption{Sensitivity to different judge models.}
        \label{fig:sensitivity_reward_model}
    \end{subfigure}
    \hfill 
    \begin{subfigure}[b]{0.48\textwidth}
        \centering
        \includegraphics[width=\textwidth]{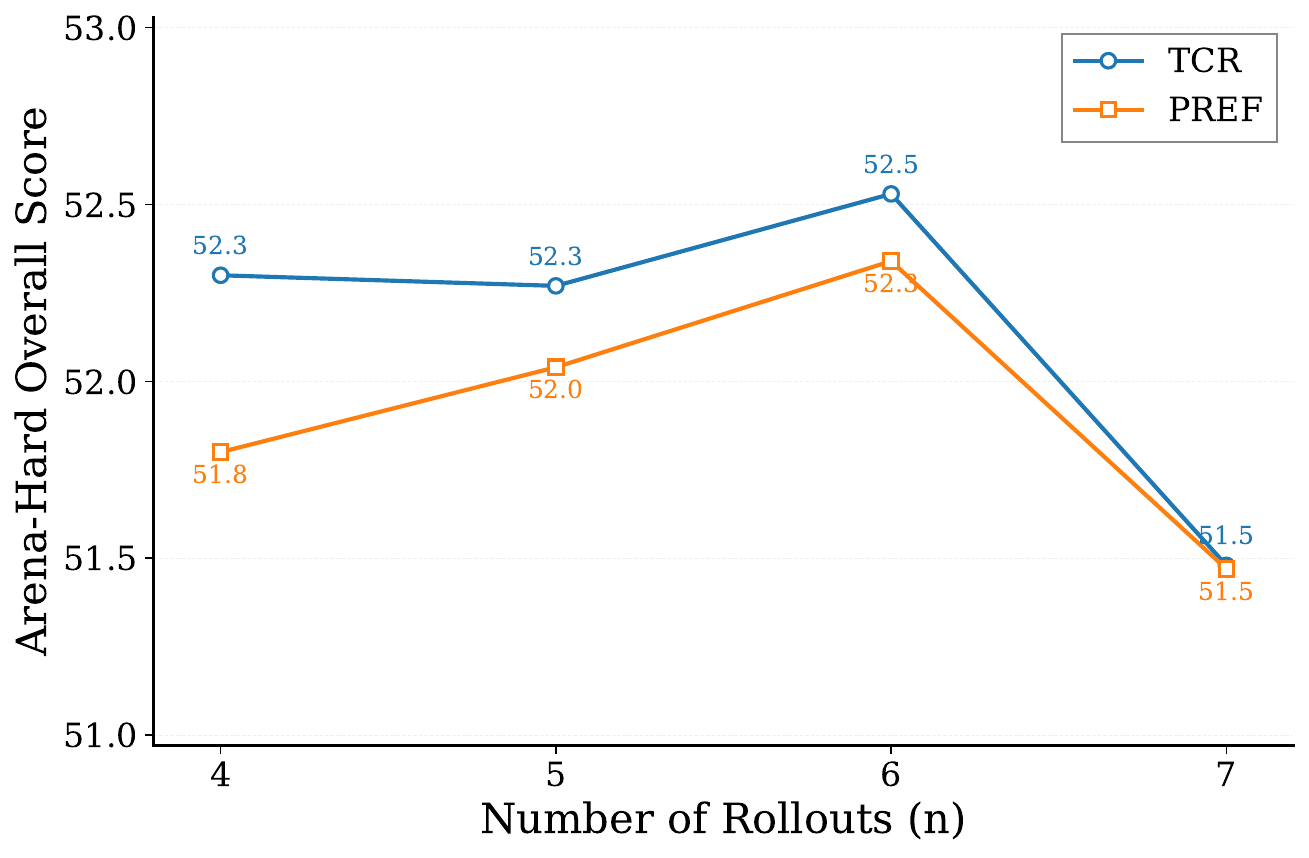}
        \caption{Sensitivity to graph size ($n$).}
        \label{fig:sensitivity_graph_size}
    \end{subfigure}
    \caption{Sensitivity analysis of TCR compared to PREF on Arena-Hard. (a) Performance across three different LLM judges. (b) Performance as the number of candidates $n$ increases from 4 to 7.}
    \label{fig:sensitivity_analysis}
\end{figure*}

\subsection{Robustness to Different Judge Models}

Using the GSPO optimizer with Qwen3-8B, Figure~\ref{fig:sensitivity_reward_model} shows that TCR's advantage is not contingent on a specific LLM judge. TCR consistently outperforms PREF across all tested judges, holding a significant average advantage on Arena-Hard. TCR exhibits far greater stability across different judges, whereas PREF's performance shows high variance. The advantage is most pronounced with less capable judges, indicating that TCR is particularly effective at extracting consensus from lower-quality judge models. This confirms that TCR is a universally applicable enhancement that robustly improves alignment regardless of the preference data source.

\subsection{Scalability with Increasing Graph Size}

Using the GSPO optimizer with Qwen3-8B, Figure~\ref{fig:sensitivity_graph_size} illustrates the performance trend as the number of candidates $n$ increases. TCR maintains a consistent performance advantage over PREF for all tested values of $n$. Both methods achieve their peak performance at $n=6$. While the advantage margin varies, TCR's ability to systematically extract consensus prevents the performance degradation one might expect from the exponential increase in potential preference cycles in a larger graph. This consistent, positive advantage demonstrates that our consensus extraction mechanism provides a reliable edge, ensuring stable and superior performance as the scale and complexity of the comparison task grow.

\end{document}